\documentclass[manuscript,screen,nonacm,acmthm=false]{acmart}

\usepackage{amsmath,graphicx,booktabs}
\usepackage{amsthm}
\usepackage[caption=false,font=footnotesize]{subfig}
\theoremstyle{acmdefinition}
\newtheorem {definition}{Definition}
\usepackage{stfloats}

\DeclareRobustCommand{\mshi}{\textcolor{black}}

\begin{document}

\title{Error Propagation in AI Medical Note Transcription: A Controlled Perturbation Study}

\author{Danica Blazanovic}
\affiliation{%
  \institution{Dept. of Computer and Electrical Engineering \& Computer Science, Florida Atlantic University}
  \city{Boca Raton}
  \state{FL}
  \country{USA}}
\email{dblazanovic2015@fau.edu}

\author{Min Shi}
\affiliation{%
  \institution{School of Computing and Informatics, University of Louisiana at Lafayette}
  \city{Lafayette}
  \state{LA}
  \country{USA}}
\email{min.shi@louisiana.edu}

\author{Xingquan Zhu}
\affiliation{%
  \institution{Dept. of Computer and Electrical Engineering \& Computer Science, Florida Atlantic University}
  \city{Boca Raton}
  \state{FL}
  \country{USA}}
\email{xzhu3@fau.edu}

\begin{abstract}
Medical note transcription \mshi{converts} medical conversations or notes into structured \mshi{fields, such as drug names, doses, and frequencies. Tools such as cTAKES, Med7, and systems based on large language models (LLMs) are used for this task to reduce labor costs and increase throughput. In practice, the input to these systems may contain errors, for example from speech recognition. How do existing artificial intelligence (AI)-based transcription systems handle such errors? Do they detect them, correct them, or propagate them downstream? This paper addresses these questions with a controlled perturbation study. Seven classes of documented medication errors are introduced, one at a time, into 7,256 medication entries from 555 MIMIC-IV discharge summaries, and the behavior of a dictionary-based system (cTAKES), a supervised tagger (Med7), and two open-weight language models (Llama and Qwen) is compared across eight configurations.} The errors are a drug-name substitution, a tenfold dose error, two frequency misreadings, and the deletion of a dose, a frequency, or a route. For each edited entry, \mshi{the output from the perturbed section is compared} with the output from the unperturbed section. An error is masked when the two outputs agree and propagated when they differ. Drug-name substitutions and tenfold dose errors \mshi{are propagated in 98.6\% to 100\% of cases in every configuration: the wrong value reaches the structured output or, less often, the field holds another value or the entry is missing from the output.} A deleted dose, frequency, or route is masked \mshi{mainly when the entry repeats the value} or the system fills in a default, so masking reflects redundancy in the note, not error detection. \mshi{Configurations that read the whole entry, the language models among them, recover more deleted values, but the language models} also move values between fields, and \mshi{under concurrent inference} their outputs change from run to run even without an edit. Extraction therefore cannot serve as a safety net for errors already in the note, and error detection must rely on a reference outside the note.
\end{abstract}




\maketitle

\section{Introduction}
\label{sec:intro}

An ambient AI scribe records a clinical encounter and drafts the clinical note for the clinician to review. In one medical group, such scribes assisted 7{,}260 physicians in more than 2.5 million encounters between October 2023 and December 2024~\cite{tierney2025ambient}. The drafted notes are not the last step, but are often used for administration and by physicians~\cite{shah2025physican}. Research study has shown that such approaches enhance clinicians' documentation-related experience and reduce burnout~\cite{olson2025use}. AI-based medical note transcription has three stages: the audio is transcribed, the transcript becomes a clinical note, and the content of the note is extracted into structured fields. The scribe performs the first two stages. This study concerns the third, for medications. Medication extraction converts the medication list of a note into fields such as drug name, dose, route, and frequency, and an error already in the note may pass into those fields. \mshi{These fields matter because other software reads them rather than the note. Medication data in free text are not accessible to computerized applications that rely on coded data~\cite{xu2010medex}. The errors are also consequential. Most confused drug-name pairs on the list of the Institute for Safe Medication Practices (ISMP) have no indication in common~\cite{cheng2018lasa}, and of 252 tenfold medication errors reported at one pediatric hospital over five years, 22 involved patient harm~\cite{doherty2012tenfold}.}

Medication errors are documented in the notes that scribes and dictation produce. In a
comparison of five scribe platforms on simulated encounters, medication-related errors were the
most common errors in the notes, as omissions and as incorrect
content~\cite{anderson2025safety}. An AI scribe evaluated on simulated medication histories
wrote wrong medication names and wrong strengths, and with audio-only input its least accurate
field was the strength and form of the medication~\cite{menz2026vision}. In dictated notes,
medication names and doses were the most common clinical content among speech recognition
errors~\cite{zhou2018errors}. One stage of propagation has been measured: for four of the five
scribe platforms, 19.5\% of the transcript errors were transmitted to the clinical
note~\cite{anderson2025safety}. The next stage has not. No study reports whether extraction
detects an error in the note, corrects it, or passes it on.

\mshi{Existing study designs cannot answer this question.} Prior work has followed naturally
occurring errors through the documentation stages \mshi{~\cite{zhou2018errors, fu2024cacer}}, so it cannot separate what an extraction
system does to an error from what it gets wrong on its own. Where errors were injected
deliberately \mshi{in studies of AI scribes}, they entered before the note was written~\cite{rabotin2026propagation,
wang2025scribe}. Controlled perturbation of the note separates the two, and
Section~\ref{sec:related} shows it has not been applied to this stage. \mshi{Where errors were injected into the text itself, the question was a different one. MEDEC
inserts errors into clinical texts and tests whether models detect and correct them when asked
to~\cite{benabacha2025medec}. Outside clinical text, perturbations of the input of language
models were used to measure the loss of accuracy under noise~\cite{moradi2021robustness}.
Neither follows an injected error from the note to the structured fields.}

This study measures that stage by controlled perturbation. It takes 7{,}256 medication entries from
555 MIMIC-IV discharge summaries and introduces one documented error at a time into every entry
eligible for the class: a drug-name substitution, a tenfold dose error, two frequency misreadings,
or the deletion of a dose, a frequency, or a route. \mshi{This gives 36{,}559 perturbed sections.} Eight
extraction configurations process each perturbed section and its unperturbed original. They are a
dictionary-based system (cTAKES~\cite{savova2010ctakes}), a supervised tagger
(Med7~\cite{kormilitzin2021med7}) reading the entry line by line or as a block, a variant of
Med7 with a transformer encoder (Med7-trf~\cite{kormilitzin2021med7,lewis2020pretrained}),
Llama~3.1 8B~\cite{llama3herd2024} under three prompts, and Qwen2.5 7B~\cite{qwen2025qwen25}
under one. The two outputs of a configuration for one perturbed section form a pair.
For the edited entry they are compared: the error is masked when they
agree and propagated when they differ, directly when the perturbed value appears in the output
and with distortion otherwise. The two outputs are also compared on the other fields of the
entry and on the other entries of the section, and a change there is interference.

\mshi{The main results are as follows. Drug-name substitutions and tenfold dose errors are masked in at most 1.4\% of scored pairs in every configuration. A deleted value is masked mainly when the entry repeats it or, for the route, when a language model supplies a default. The language models also change other fields of the edited entry, and both they and the transformer-based tagger change other entries of the section.} To conclude, our main contributions include:
\begin{enumerate}
\item[\mshi{(1)}] A per-class measurement of error propagation from the medication list
of a clinical note to the structured medication output, on 36{,}559 perturbations;
\item[\mshi{(2)}] Evidence
that masking arises from information outside the edited span, a repeated value in the entry or a
default, and not from error detection;
\item[\mshi{(3)}] A comparison of the three categories of extraction
systems, and of prompts, on the same errors, including interference with other fields of the
edited entry and with other entries; and
\item[\mshi{(4)}] A comparison of seven drug-name similarity measures
by how well they recover the documented confusion partner on the ISMP list.
\end{enumerate}

Section~\ref{sec:related} reviews related work, Section~\ref{sec:methods} \mshi{presents the methodology},
Section~\ref{sec:setup} \mshi{the experimental setup}, and Section~\ref{sec:results} the results.
Section~\ref{sec:discussion} discusses them, and Section~\ref{sec:conclusion} concludes.
For ease of understanding, Table~\ref{tab:terms} lists the terms and abbreviations used in the paper.

\section{Related Work}
\label{sec:related}

\subsection{Controlled error injection in clinical text}

The closest precedent to this study is MEDEC, a benchmark of 3,848 clinical
texts~\cite{benabacha2025medec}. Annotators inserted at most one error per text by swapping a
concept with a similar alternative, across five categories: diagnosis, management, treatment,
pharmacotherapy, and causal organism. About half the texts are error-free controls. MEDEC tests
whether models explicitly detect and correct errors; this study examines how extraction systems
process the error when nothing asks them to check it. The perturbation classes of this study derive
from documented medication-safety sources, the ISMP List of Confused Drug Names and the ISMP
List of Error-Prone Abbreviations, Symbols, and Dose
Designations~\cite{ismp2024confused, ismp2024abbreviations}, and from the omission errors observed
in the output of an AI scribe~\cite{menz2026vision}, rather than \mshi{from} annotator-crafted
contradictions as in MEDEC~\cite{benabacha2025medec}. Outside clinical text, Moradi and Samwald
applied character-level and word-level perturbations to the inputs of four neural language
models on five tasks, including named-entity recognition, and the scores of the models
decreased~\cite{moradi2021robustness}. That study measures the loss of accuracy under noise.
This study follows each perturbation to the structured output.

The perturbations reproduce documented medication errors. Cheng et al. analyzed 278
look-alike, sound-alike (LASA) drug pairs from the ISMP list and found that 59\% \mshi{have no overlap in indications}, confirming that LASA substitutions replace one clinical entity with another
rather than with a spelling variant~\cite{cheng2018lasa}. At the documentation level, Menz et
al. categorized the errors of an AI scribe as omission or commission errors on 100 simulated
medication histories~\cite{menz2026vision}. The commission errors included wrong medication
names and wrong strengths.

\subsection{Error propagation across documentation stages}

Prior work has \mshi{followed errors} across multi-stage clinical workflows. Zhou et al. tracked
217 dictated notes through speech recognition, transcriptionist editing, and physician sign-off,
recording which errors survived each stage~\cite{zhou2018errors}. CACER \mshi{showed} how
automated event-extraction errors cascade into relation extraction across BERT, Flan-T5,
Llama~3, and GPT-4~\cite{fu2024cacer}. Both studies followed errors that arose on their own, in
dictation or in a preceding model, so what a stage does to an error cannot be separated from the
errors that the stage makes on its own. Controlled error injection avoids this problem: the fate
of a known error is tracked separately from the baseline performance of the system. Two studies
of AI scribes inject errors before the note is written. Rabotin et al. scripted 20 interpreter
errors into five simulated encounters, and two scribes carried 55\% and 60\% of them into the
clinical note~\cite{rabotin2026propagation}. Wang et al. masked 10\% of the words of the
transcripts, or altered 10\% of the speaker labels, and measured the quality of the generated
note~\cite{wang2025scribe}. Neither study reaches the stage from the note to the structured
fields.

\begin{table*}[!t]
\caption{Terms and abbreviations used in this study. Each row gives a term and its meaning. The
upper part lists the terms defined in this study; the lower part lists abbreviations and the two
similarity scores.}
\label{tab:terms}
\centering
\footnotesize
\setlength{\tabcolsep}{4pt}
\begin{tabular}{p{0.20\textwidth}p{0.76\textwidth}}
\toprule
Term & Meaning \\
\midrule
Discharge summary & The clinical note; one per hospital admission \\
Section & A medication list in the note, from its header to the next header \\
Entry & One medication with its fields: a numbered line and, in the \emph{RX}-line template, the pharmacy line beneath it \\
Field & Name, dose, route, or frequency of an entry \\
Span & The character positions of a field in the note \\
Entry block & The entry's lines, from its numbered line to the next numbered or blank line \\
Perturbation target & An entry that meets the five inclusion criteria
  (Section~\ref{sec:preparation}). It receives one error for each class it is eligible for \\
Perturbation & One edit of one field of one entry; the error that this study introduces into the note \\
Perturbed value & The value that a perturbation writes into the field; for a deletion, the absence of the value \\
\emph{Sig:} template & The entry has no \emph{RX} line; its numbered line can carry a \emph{Sig:} instruction \\
\emph{RX}-line template & The numbered line is followed by a pharmacy \emph{RX} line that can repeat the drug and its fields \\
Copy & A second occurrence of a deleted value in the entry block \\
System & One of the four extraction tools: cTAKES, Med7, Llama, or Qwen \\
Configuration & One system under one setting (Table~\ref{tab:configs}) \\
Pipeline & The function from a section to field values that a configuration computes; written
  $\mathcal{T}(\cdot)$ in the definitions \\
Reference output & A configuration's output for the unperturbed section \\
Perturbed output & Its output for the section with one perturbation \\
Pair & The perturbed output and the reference output of the same section, compared for one
  perturbation \\
Scored pair & A pair whose reference output extracts the entry correctly: the name and the field
  that the class edits match the label \\
Scored entry & The edited entry of a scored pair \\
Fate & Outcome of the perturbation on a scored pair: M, P, or D \\
Masked (M) & The edited field is the same in both outputs \\
Propagated directly (P) & The edited field of the perturbed output holds the perturbed value
  (Eq.~\eqref{eq:fate}) \\
Propagated with distortion (D) & The edited field holds a third value, or the edited entry is
  missing from the perturbed output \\
Masking rate, masked share & The share of a configuration's scored pairs of a class that are masked \\
Exact criterion & P requires that the field equals the perturbed value \\
Token-level criterion & P requires that the field contains the perturbed value and no other new word; the primary criterion \\
Distortion type & Origin of the distorted value: missing, another field of the entry, the entry's text, another entry, or produced  \\
Intra-entry interference & A field of the edited entry other than the edited one changed \\
Inter-entry interference & An entry of the section other than the edited one changed \\
Repeat runs & Further runs of a language-model configuration on the unperturbed section, with identical settings \\
Total error & Number of fields changed by one perturbation; entry level counts the edited entry, section level the whole section \\
Easy stratum & Entries every configuration extracts correctly from the unperturbed section \\
Hard stratum & Entries no configuration extracts correctly from the unperturbed section \\
Intermediate stratum & Entries some configurations extract correctly, and others do not \\
\midrule
PO & By mouth \\
QD, daily & Once a day \\
BID, TID, QID & Two, three, four times a day \\
QHS & At bedtime \\
PRN & As needed \\
Q1H & Every hour \\
SC, SL & Subcutaneous; sublingual \\
LASA & Look-alike, sound-alike drug names \\
ISMP & Institute for Safe Medication Practices \\
POCA & Phonetic and Orthographic Computer Analysis, the FDA's program for scoring the similarity of drug names \\
Composite score & The name-similarity score computed in this study, modeled on POCA (Section~\ref{sec:classes}) \\
\bottomrule
\end{tabular}
\end{table*}

\subsection{Measuring drug-name similarity}

Similarity between drug names predicts confusion errors. Lambert et al.
showed this on 1{,}127 confused pairs with 22 orthographic and phonetic
measures~\cite{lambert1999similarity}. Kondrak and Dorr then asked which
measure works best~\cite{kondrak2006confusable}. They used BI-SIM for
spelling and ALINE for sound; ALINE compares phonetic transcriptions of the
names. In their evaluation, each name is compared with all other names, and
a measure is scored by the share of the name's known confusion partners
among the most similar names. Combined measures scored highest. \mshi{The Phonetic and Orthographic Computer Analysis (POCA) program of the U.S. Food and Drug Administration (FDA)} combines measures of both kinds. As Emmerton et al. describe
it, an orthographic score, the mean of BI-SIM and normalized edit
similarity, is averaged with ALINE~\cite{emmerton2020lasa}. The FDA revised
the orthographic part in December 2016, and its naming guidance sets 70\%
and 55\% of the combined score as the thresholds for high and moderate
similarity~\cite{fda2020proprietarynames}. Later studies scored known ISMP
pairs with POCA descriptively~\cite{karet2023lasa}. The ranking evaluation
itself was continued on the United States Pharmacopeia (USP) confused-name
collection~\cite{millan2019softbisim}. To our knowledge, no peer-reviewed study since 2006 treats the ISMP list as the ground truth and compares measures by the rank at which the documented confusion partner is recovered; the search covered PubMed, the ACL Anthology, IEEE Xplore, and Google Scholar through September 2026.

\subsection{Prompting language models for extraction}

Chain-of-thought prompting asks the model to reason before it answers~\cite{wei2022cot}. Kojima
et al. showed that one instruction to reason step by step achieves this without
examples~\cite{kojima2022zeroshot}. For entity extraction, PromptNER asks the model to justify
each candidate entity against the text before labeling it~\cite{ashok2023promptner}. Reasoning
does not always help extraction. Nagar et al. evaluated biomedical named-entity recognition and
classification and found standard prompting more accurate than
chain-of-thought~\cite{nagar2025llms}. The three prompts of this study, zero-shot, one-shot, and
chain-of-thought with justification, test whether reasoning changes what the model does with a perturbation.

\subsection{Metamorphic testing and run-to-run variation}

A test normally compares the output of a program with the expected
output. Metamorphic testing is used when the expected output is hard
to obtain~\cite{segura2016survey}. A source input is transformed into
a follow-up input, and a metamorphic relation states how the two
outputs must relate. A violated relation reveals a fault, although
the correct output of neither input is known. When the output of a
program differs from execution to execution, one comparison cannot
decide whether the relation holds. Guderlei and Mayer therefore
generate several outputs and test the relation with statistical
hypothesis tests~\cite{guderlei2007statistical}. Language models show
such variation. At{\i}l et al. ran five hosted models ten times on
eight tasks under settings expected to be deterministic, and no model
returned identical outputs across the runs on every
task~\cite{atil2025nondeterminism}. Pu et al. separated this variation
from the effect under study. They submitted 50 patient questions
three times through each access mode of a model. They then compared
the overlap of response content across modes with the overlap among
repeated runs in the same mode~\cite{pu2026auditing}. This study
applies the same design to the fields that a perturbation does
not touch. The unperturbed section is the source input and the
perturbed section is the follow-up input. The relation is that every
field other than the edited one stays unchanged, in the edited entry
and in the other entries. For the language models, the perturbed
output is compared with repeat runs of the unperturbed section.

\section{\mshi{Methodology}}
\label{sec:methods}
\begin{figure*}[t]
\centering
\includegraphics[width=\textwidth]{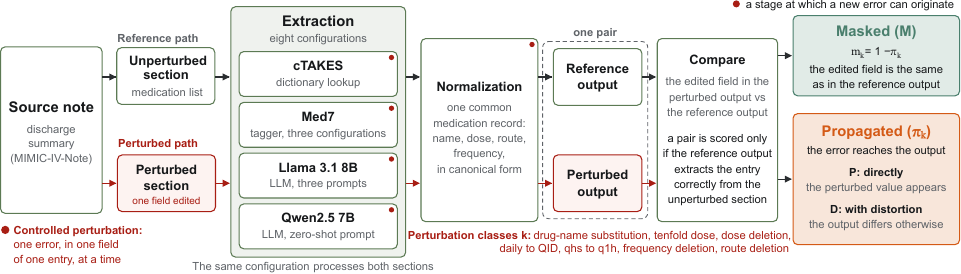}
\Description{A flow diagram from left to right with two parallel paths that start at the source
note. On the upper path, the unperturbed section passes through extraction and normalization and
gives the reference output. On the lower path, the section with one perturbation passes through
the same extraction and normalization and gives the perturbed output. Both outputs enter a
comparison box, which leads to two outcomes: masked, or propagated, directly or with distortion.}
\caption{The framework of the single-error perturbation/intervention analysis. The same configuration processes a medication section
twice. On the reference path (upper) it processes the section as written, which gives the
reference output. On the perturbed path (lower) it processes the same section with one field of one
entry edited and the rest held fixed, which gives the perturbed output. Each configuration
processes an unperturbed section once, and that output is the reference for every perturbed copy
of the section\mshi{; the language-model configurations repeat this run to measure run-to-run variation (Section~\ref{sec:measurement})}. Both outputs are normalized. They form one pair (dashed frame), and the edited entry is compared between them. The error is \emph{masked} when the edited field is the same in both outputs and \emph{propagated} when it differs, directly if the perturbed value appears in the output and with distortion otherwise. Only entries the configuration extracts correctly from the unperturbed section are
scored. Red dots mark stages at which a new error can originate; $\pi_k$ is the propagation
\mshi{rate} for perturbation class $k$, and $m_k = 1 - \pi_k$ is the masking rate.}
\label{fig:design}
\end{figure*}

This study measures error propagation through one stage of the documentation pipeline, from the
medication list of a clinical note to the structured medication output. An error of known class
and location is introduced into one entry of a medication section. The same configuration
processes the section with and without the error, and the fate of the error is read from the two
outputs (Fig.~\ref{fig:design}). The procedure \mshi{is repeated for} eight extraction configurations \mshi{(Section~\ref{sec:configurations})}, so
the results \mshi{can be compared} across configurations as well as across perturbation classes.

\subsection{\mshi{Units of analysis}}
\label{sec:data}
This study uses the discharge summary, the clinical note written at the end of
a hospital admission. It covers the reason for admission, the hospital
course, and the discharge instructions, including the medication lists.

\begin{figure}[!t]
\centering
\includegraphics[width=\textwidth]{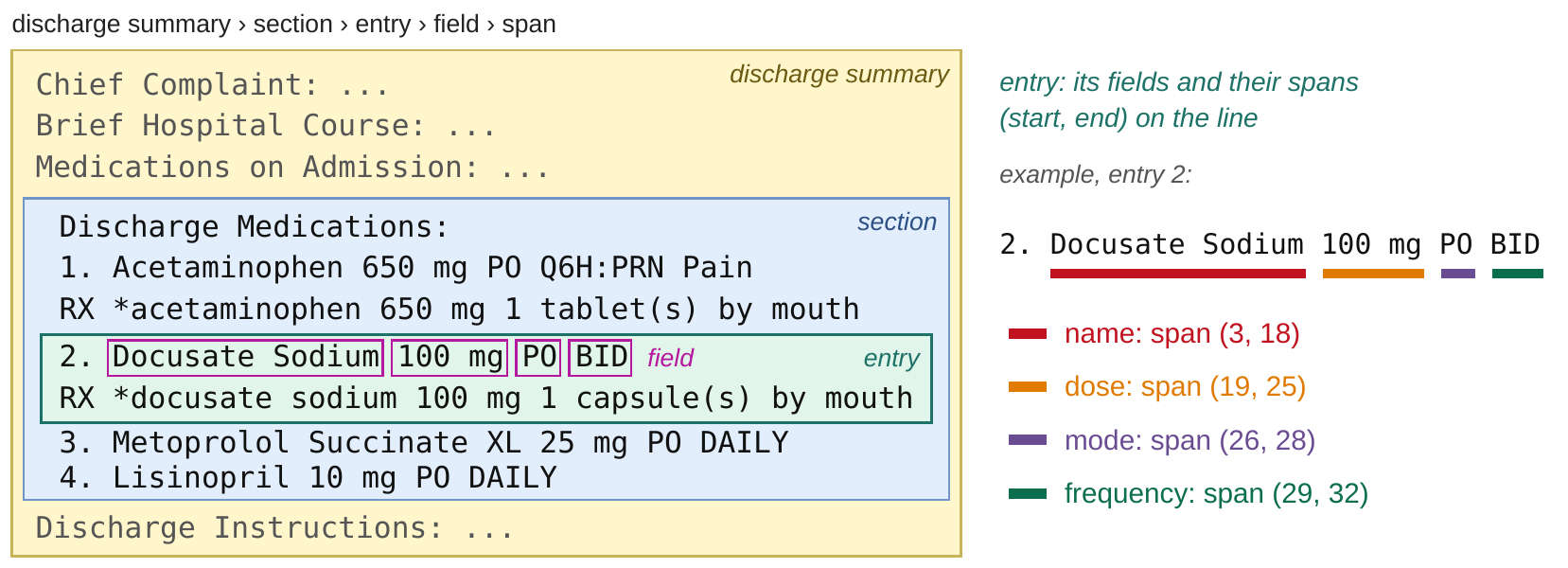}
\Description{On the left, a constructed discharge summary drawn as nested boxes: the summary
contains a Discharge Medications section with four numbered medication entries, entry 2 is
outlined with its pharmacy RX line, and its four fields are boxed. On the right, entry 2 is
repeated with its name, dose, mode, and frequency underlined and the character span of each.}
\caption{The units of the study. Left: the discharge summary (yellow) contains sections; the
Discharge Medications section (blue) contains medication entries; entry~2 (green) consists of a
numbered line and its pharmacy \emph{RX} line. Right: entry~2 with its four fields underlined and
the character spans that locate them, counted here within the line; this study calls the \emph{mode} field the route. The text is a constructed example in the format of the
notes, because the MIMIC-IV text may not be reproduced under its data use agreement.}
\label{fig:units}
\end{figure}

The units of the study, from largest to smallest, are the note, its
sections, the medication entries in a section, the fields of an entry, and
the spans that locate the fields (Fig.~\ref{fig:units}). A section runs from
one top-level header of the note to the next; the medication sections are
Discharge Medications and Medications on Admission. Following the i2b2
schema~\cite{uzuner2010medication}, a medication entry is one mention of a
medication together with its fields: name, dose, mode, frequency, duration,
and reason. This study uses four of them: name, dose, mode, and frequency.
\mshi{It} calls the mode field the route; the dose field is the
amount written with the name, such as 10~mg. Each field is a span, a start
and end character position in the note. The label set \mshi{(Section~\ref{sec:sources})} marks the spans and
links the spans of one entry by a group code.

The medication sections follow two templates. In the \emph{RX}-line template, the numbered
line of an entry is followed by a pharmacy \emph{RX} line that can repeat the drug and its
fields. In the \emph{Sig:} template, the entry has no \emph{RX} line, and its numbered line
can carry a \emph{Sig:} instruction. The entry block is
the entry's lines, from its numbered line to the next numbered or blank
line. In the \emph{RX}-line template the block can hold a second copy of a
value, so a value deleted from the numbered line may remain in the block. An entry is counted in
the \emph{RX}-line template when its block holds an \emph{RX} line, and in the \emph{Sig:}
template otherwise.

\subsection{Perturbation types}
\label{sec:classes}

\begin{figure*}[!t]
\centering
\includegraphics[width=0.7\textwidth]{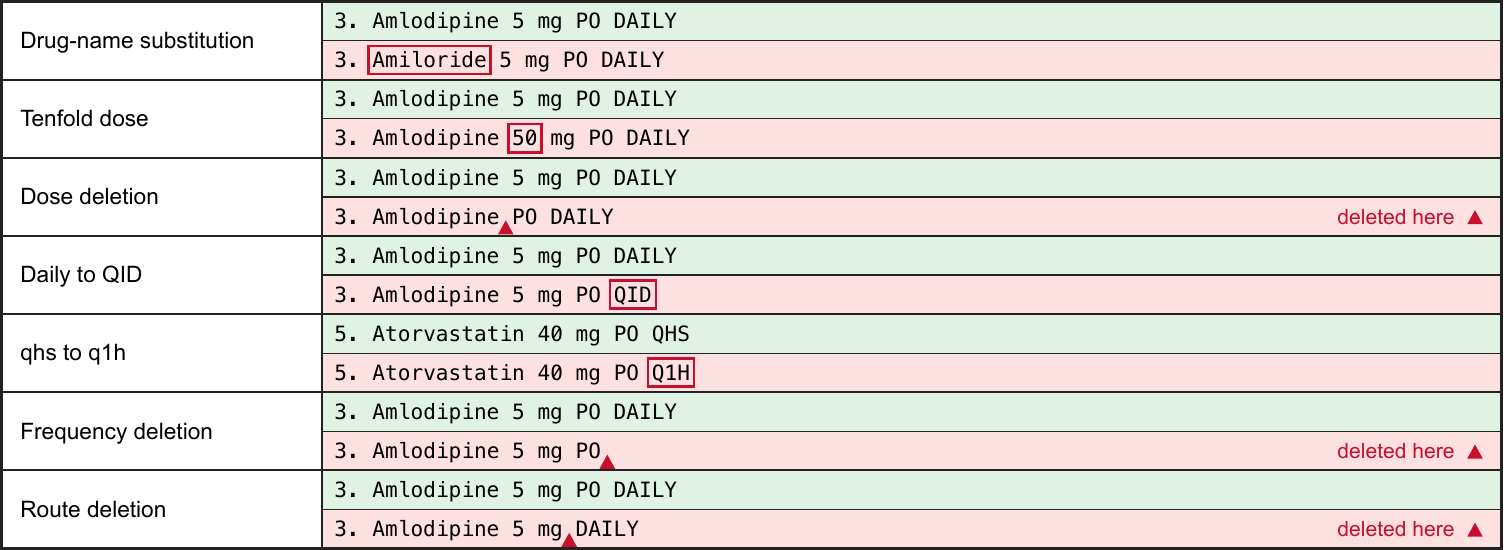}
\Description{Fourteen constructed medication entries, two for each perturbation class. For
each class the upper line is the unperturbed entry and the lower line its perturbed copy, with the
edit marked: a substituted drug name, a tenfold dose, a deleted dose, daily changed to QID, QHS
changed to Q1H, a deleted frequency, and a deleted route.}
\caption{Examples of seven perturbation classes on constructed entries. Each row
shows the unperturbed entry (green) and its perturbed copy (pink); the edited
span is boxed in red, and a deleted span is marked at its position.
Amlodipine and amiloride are an ISMP confused pair.}
\label{fig:injections}
\end{figure*}

A perturbation edits one field of one entry and leaves the rest of the
section unchanged. Because the fields are spans, the class and the location
of the perturbation are known. Seven classes are applied, \mshi{by} four
mechanisms: a drug-name substitution, a tenfold dose error, the deletion of
the dose, the frequency, or the route, and two frequency misreadings. Each
reproduces a documented medication error. Fig.~\ref{fig:injections} shows one
constructed example of each.

\subsubsection{Drug-name substitution}
A drug-name substitution replaces the name of the drug with the name of another drug. It
reproduces the error type \emph{wrong drug} of the taxonomy of the National Coordinating Council
for Medication Error Reporting and Prevention (NCC MERP), which lists name confusion among its
causes~\cite{nccmerp1998taxonomy}. The name is replaced by a confusable name from one of three
tiers, each used only when the previous one yields no candidate:
\begin{enumerate}
\item the confused partner from the ISMP List of Confused Drug
  Names~\cite{ismp2024confused}, matched on the name as written, brand or
  generic;
\item the most similar name under a composite score modeled on the FDA's POCA score, at 0.70 or above;
\item the most similar name at 0.55 to 0.69.
\end{enumerate}

The composite score of tiers~2 and~3 is computed in this study. It has the form that Emmerton et al. describe for the FDA's POCA program~\cite{emmerton2020lasa}: an orthographic similarity, the mean of BI-SIM and normalized edit similarity, is averaged with the ALINE phonetic similarity~\cite{kondrak2006confusable}. ALINE is applied here to the name as spelled, not to a phonetic transcription. The score ranges from 0 to 1. The cut points 0.70 and 0.55 are taken from the FDA's naming guidance, which defines them for the POCA score as the thresholds for high and moderate similarity~\cite{fda2020proprietarynames}.

The candidates of tiers~2 and~3 are drug names in clinical use,
matched brand for brand and ingredient for ingredient. A candidate
with the same concept as the original in RxNorm, the drug vocabulary
of the U.S. National Library of
Medicine~\cite{nelson2011rxnorm}, is excluded. Names that consist
only of salt or electrolyte words, such as \emph{sodium chloride},
receive no substitute from these tiers and are not candidates. In all
three tiers, a name is not used as a substitute when it may be a
variant of the original drug, that is, when:

\begin{itemize}
\item one name begins with the other;
\item the two names share a word; or
\item a word of one name and a word of the other have the same first six letters or differ by at most two single-letter edits.
\end{itemize}

One substitute is then selected. In tier~1, a name with several remaining ISMP partners receives
the first in alphabetical order. In tiers~2 and~3, the 50 candidates with the highest orthographic
similarity are scored with the composite, and the one with the highest score is the substitute; a
tie goes to the first name in alphabetical order. The replacement keeps the capitalization of the
original. Section~\ref{sec:similarity} compares the composite with other similarity measures and
with the FDA's tool.

\subsubsection{Tenfold dose error}

The first numeric value in the dose span is changed by a factor of ten.
The ISMP List of Error-Prone Abbreviations, Symbols, and Dose
Designations~\cite{ismp2024abbreviations} identifies two errors that
can cause a tenfold overdose: reading 1.0 mg as 10 mg because of a
trailing zero, and reading .5 mg as 5 mg because of a missing leading
zero. Rather than reproducing these specific errors, the perturbation
changes the numeric value directly. If the value is a whole number
of at least 10 ending in zero, it is divided by ten (20 mg becomes
2 mg). Otherwise, it is multiplied by ten (5 mg becomes 50 mg).
The unit remains unchanged.

\subsubsection{Field deletion}

The dose, frequency, or route span is deleted, so the medication appears without its dose,
frequency, or route. These are three classes with the same mechanism. They reproduce the omission
errors that Menz et al. observed in the medication histories written by an AI
scribe~\cite{menz2026vision}. With audio-only input, their scribe made 358 omission errors, and
the field with the lowest accuracy was the strength and form of the medication (28\%). The
strength is the field this study calls the dose.

\subsubsection{Frequency misreadings}
Two classes change the frequency. In the first, an entry taken once a day has its frequency replaced by \emph{qid}, four times a day, whatever the original wording (\emph{daily}, \emph{once a day}, \emph{qd}). In the second, an entry taken at bedtime, written \emph{qhs}, has its frequency replaced by \emph{q1h}, every hour. Both are misreadings on the ISMP list of error-prone abbreviations~\cite{ismp2024abbreviations}: QD read as QID, and qhs read as qhr, every hour, which these notes write as q1h (Section~\ref{sec:injection}). The replacement keeps the capitalization of the original.

\subsection{Propagation types}
\label{sec:types}
Let $x$ be a medication section, $e \in x$ an \emph{entry} of $x$, and
$f \in e$ a \emph{field} of $e$. $x[e\!:\!f]$ denotes field $f$ of entry
$e$ in $x$. A medical note transcription pipeline $\mathcal{T}(\cdot)$ is a function from a section to the values of its fields. $\mathcal{T}(x[e\!:\!f])$ is the value the pipeline returns for field $f$ of entry $e$ when it processes $x$, or $\varnothing$ when it returns nothing for that field. Similarly, $\mathcal{T}(x[e\!:])$ \mshi{is the set of} all extracted fields of entry $e$ \mshi{in} section $x$, and $\mathcal{T}(x[:])$ denotes the extracted fields of all entries \mshi{in} section $x$. \mshi{A set difference excludes a member.} For example, $\mathcal{T}(x[:]) \setminus \mathcal{T}(x[e\!:\!f])$ denotes the set of all extracted fields from $x$, except the one corresponding to the field $f$. Values are compared after normalization, which Section~\ref{sec:measurement} describes.

\begin{definition}[Reference output]
The \emph{reference output} is \mbox{$\mathcal{T}(x[:])$}, the pipeline's output on the section
as written. The entry is \emph{scored} when the reference output agrees with the label in the
name and in field $f$. A dose agrees when its number and unit equal those of the label. A
frequency or a route agrees when it contains every word of the labeled value. A name agrees when
it contains every word of the labeled name, or the labeled name contains every word of it.
\end{definition}

A perturbation replaces one field inside an entry. $\tilde{f}$
denotes a perturbed value of $f$, produced by one of the perturbation
types above; for a deletion, $\tilde{f} = \varnothing$.
$\tilde{x}$ denotes the section in which the
span of $f$ holds $\tilde{f}$ and nothing else has changed. \mshi{This study injects a single field perturbation into a section $x$, so} $\tilde{x}[:]$, $\tilde{x}[\tilde{e}\!:]$ and
$\tilde{x}[\tilde{e}\!:\!\tilde{f}]$ each denote the perturbed
section, the edited entry, and the edited field, respectively.
Accordingly, $\mathcal{T}(\tilde{x}[\tilde{e}\!:\!\tilde{f}])$
denotes the value the extraction pipeline returns for the field $f$
when it processes the perturbed section. The extracted fields
corresponding to entry $\tilde{e}$ are denoted by
$\mathcal{T}(\tilde{x}[\tilde{e}\!:])$, and all fields extracted from
section $\tilde{x}$ are denoted by $\mathcal{T}(\tilde{x}[:])$, the
\emph{perturbed output}.

On a scored entry, the perturbation has one of three fates, given by
Eq.~\eqref{eq:fate}. The first case that holds applies.
\begin{equation}
\mathrm{Fate}(\tilde{e}) =
\begin{cases}
\mathrm{M} & \text{if } \mathcal{T}(\tilde{x}[\tilde{e}\!:\!\tilde{f}]) = \mathcal{T}(x[e\!:\!f]),\\[2pt]
\mathrm{P} & \text{if } \mathcal{T}(\tilde{x}[\tilde{e}\!:\!\tilde{f}]) \simeq \tilde{f},\\[2pt]
\mathrm{D} & \text{otherwise.}
\end{cases}
\label{eq:fate}
\end{equation}
The error is \emph{masked} (M) when the configuration returns the same value for the field with
and without the perturbation. It is \emph{propagated directly} (P) when the returned value is
the perturbed value, and \emph{propagated with distortion} (D) in every other case: the field
holds a third value, or entry $e$ is missing from the perturbed output. The error is
\emph{propagated} in cases P and D. The relation $\simeq$ in the second case is defined in two
ways, which give two criteria.

\begin{definition}[Exact criterion]
$\mathcal{T}(\tilde{x}[\tilde{e}\!:\!\tilde{f}]) \simeq \tilde{f}$ if and only if the entry is
returned and $\mathcal{T}(\tilde{x}[\tilde{e}\!:\!\tilde{f}]) = \tilde{f}$. For a deletion,
$\tilde{f} = \varnothing$, so the field must be empty.
\end{definition}

\begin{definition}[Token-level criterion]
Let $W(\cdot)$ denote the set of words of a value.
$\mathcal{T}(\tilde{x}[\tilde{e}\!:\!\tilde{f}]) \simeq \tilde{f}$ if and only if the entry is
returned and either $\mathcal{T}(\tilde{x}[\tilde{e}\!:\!\tilde{f}]) = \tilde{f}$ or
\begin{align*}
W(\tilde{f}) &\subseteq W\bigl(\mathcal{T}(\tilde{x}[\tilde{e}\!:\!\tilde{f}])\bigr) \\
 &\subseteq \bigl(W(\mathcal{T}(x[e\!:\!f])) \setminus W(f)\bigr) \cup W(\tilde{f}).
\end{align*}
The field then holds every word of the perturbed value, and every other word in it was already in
the field in the reference output. For a deletion, $W(\tilde{f}) = \varnothing$: no word of the deleted value is in the field, and no new word has appeared. The words are taken from the value
as returned, in lower case and without punctuation. For a drug name, text in brackets or
parentheses and salt and form words are also removed.
\end{definition}

The two criteria differ when the field holds the perturbed value together with text that it
already held in the reference output. For example, the frequency \emph{Q6H} is deleted from an entry
whose frequency field holds \emph{Q6H PRN} (every six hours, as needed) in the reference output,
and the same field holds \emph{PRN} in the perturbed output.
The field is not empty, so the error is distorted under the exact criterion. The deleted value
is gone and no new word has appeared, so it is direct under the token-level criterion. For the
two dose classes the value is a number and a unit, compared together, and the two criteria
give the same fate. The criteria are analogous to the phrase-level and token-level scores of the
i2b2 medication challenge~\cite{uzuner2010medication} and to the strict and lenient matching of
the n2c2 challenge~\cite{henry2020n2c2}. All results use the token-level criterion, and
Section~\ref{sec:results} reports how the exact criterion differs.

\begin{definition}[Masking rate and propagation \mshi{rate}]
For a perturbation class $k$ and a pipeline with $n$ scored entries of that
class, the \emph{masking rate} is $m_k = |\mathrm{M}|/n$ and the
\emph{propagation \mshi{rate}} is $\pi_k = (|\mathrm{P}| + |\mathrm{D}|)/n
= 1 - m_k$.
\end{definition}

The terms \emph{propagated} and \emph{masked} come from the fault-tolerance
taxonomy~\cite{avizienis2004dependability}. An error is propagated when it reaches the output of
a component, where it becomes a fault for the next component. An error is masked when it does
not reach the output. In that taxonomy, masking is deliberate: a component is built to
compensate for the error. In this study it is not: \mshi{no configuration is built or prompted to compensate for an error in the note, and Section~\ref{sec:results} shows where the masking comes from.}

The fate classification of Eq.~\eqref{eq:fate} considers only the edited
field. The other fields of the same entry and the other entries of the
section are \mshi{also} compared between the reference and the perturbed output.

\begin{definition}[Intra-entry interference]
Let the edited entry be returned in the perturbed output. The edit
causes \emph{intra-entry interference} when
$\mathcal{T}(\tilde{x}[\tilde{e}\!:]) \setminus
\mathcal{T}(\tilde{x}[\tilde{e}\!:\!\tilde{f}])
\neq \mathcal{T}(x[e\!:]) \setminus \mathcal{T}(x[e\!:\!f])$, that
is, when a field of the edited entry other than the edited one
changed. An edited entry that is absent from the perturbed output is
not counted as intra-entry interference. Its absence is the way the
perturbation propagated, and it is reported as propagation with
distortion. The intra-entry interference rate of a class is the share
of its scored entries with such a change, among the scored entries
returned in the perturbed output.
\end{definition}

\begin{definition}[Inter-entry interference]
The edit causes \emph{inter-entry interference} when
$\mathcal{T}({\tilde{x}[:]}) \setminus \mathcal{T}({\tilde{x}[\tilde{e}\!:]})
\neq \mathcal{T}({x[:]}) \setminus \mathcal{T}({x[e\!:]})$, that is, when an
entry other than the edited one changed. The inter-entry interference rate
of a class is the share of its scored entries with such a change.
\end{definition}

\begin{definition}[Total error]
The \emph{total error} of a scored entry is the number of fields
whose value differs between the perturbed output and the reference
output. It is the sum of three parts. The first part, $a$, is the
edited field $f$: $a = 1$ when the error propagated and $a = 0$ when
it was masked. The second part, $b$, is the number of the other three
fields of $e$ that changed. The third part, $c$, is the number of
fields that changed in the other entries of the section. At the level
of the entry the total error is
\begin{equation}
t_e = a + b = \bigl|\{\, g \in e :
  \mathcal{T}(\tilde{x}[\tilde{e}\!:\!g]) \neq
  \mathcal{T}(x[e\!:\!g]) \,\}\bigr| ,
\label{eq:total-entry}
\end{equation}
and at the level of the section it is
\begin{equation}
\begin{aligned}
t_x &= a + b + c, \\
c &= \sum_{e' \neq e} \bigl|\{g \in e' :
  \mathcal{T}(\tilde{x}[e'\!:\!g]) \neq
  \mathcal{T}(x[e'\!:\!g])\}\bigr| .
\end{aligned}
\label{eq:total-section}
\end{equation}
The sum in $c$ runs over every other entry $e'$ that the pipeline
returns in either output, with or without a label. An entry present
in one output and absent from the other contributes all four of its
fields. This holds for the edited entry as well: when it is absent
from the perturbed output, $a = 1$ and $b = 3$. Each field of the
section appears in one part only, so no field is counted twice. The
mean of $a$ over the scored entries of a class is the propagation
\mshi{rate} $\pi_k$. The edit can change only the output of its own
section, because each section is processed on its own.
\end{definition}

The two interference measures are invariance tests in the sense of behavioral testing: an input
change that should not affect a part of the output must leave that part
unchanged~\cite{ribeiro2020checklist}. The interference rates and the total error are computed
from the same comparison of fields. Intra-entry interference occurs when $b > 0$ and the edited
entry is returned. Inter-entry interference occurs when $c > 0$. Each rate is the share of
scored pairs in which the interference occurs\mshi{; for intra-entry interference the share is taken among the scored pairs whose edited entry is returned}. The total error is the number of changed fields,
reported as its mean over the scored pairs.

\begin{definition}[Distortion types]
\label{def:kinds}
On a distorted entry, let $v = \mathcal{T}(\tilde{x}[\tilde{e}\!:\!\tilde{f}])$.
The type of distortion is the first of the following that holds:
\begin{enumerate}
\item \emph{missing}: $v = \varnothing$, the field or the entry is not
  returned;
\item \emph{other field}: $v \in \mathcal{T}(x[e\!:]) \setminus \mathcal{T}(x[e\!:\!f])$,
  the value of another field of the same entry;
\item \emph{entry text}: $v$ occurs in the block of $\tilde{e}$ outside its
  labeled fields;
\item \emph{other entry}: $v$ occurs in $\tilde{x}$ outside the block of
  $\tilde{e}$;
\item \emph{produced}: $v$ occurs nowhere in $\tilde{x}$.
\end{enumerate}
\end{definition}

\section{Experimental Setup}
\label{sec:setup}

\subsection{Data sources}
\label{sec:sources}
The notes come from MIMIC-IV-Note~v2.2~\cite{johnson2023mimicnote}. It holds
331{,}794 deidentified discharge summaries from 145{,}915 patients at the
Beth Israel Deaconess Medical Center. Protected health information was
removed under the HIPAA Safe Harbor provision, by \mshi{the union of a rule-based method and} a neural deidentification model, and each removed span was
replaced with three underscores. The hospital's Institutional Review Board
reviewed the collection of the data, waived informed consent, and approved
the sharing of the data for research~\cite{johnson2023mimicnote}.

The medication labels come from a set covering 600 of these
summaries~\cite{goel2023medlabels}. One file contains no annotations, which
leaves 599 notes carrying 64{,}106 spans in 23{,}778 entries. The labels were
generated by the dataset's authors with a large language model pipeline,
not by human annotators, and they contain systematic errors. Laboratory
values are tagged as medications (a potassium level \mshi{is tagged} like a potassium order),
allergy lists are tagged as single spans covering many drug names, and dose
fragments such as ``500~mg'' are tagged as names.

Besides the notes and their labels, the study draws on two auxiliary tables and one drug
terminology:

\begin{itemize}
    \item \textbf{The structured record:} MIMIC-IV on FHIR~v2.1 supplies the structured record of the same admissions in the Fast
Healthcare Interoperability Resources (FHIR) format~\cite{bennett2024mimicfhir,
bennett2023fhirpaper, johnson2023mimiciv, johnson2023mimicived}. The record serves one purpose:
its drug names contribute to the vocabulary used to select substitute names. It is not a scoring
reference. The notes and the record connect only at the level of the admission, and one
admission carries a median of 30 medication requests, so an entry of a note has no identified
counterpart in the record.
    \item \textbf{The admission diagnosis:} The primary diagnosis of each admission comes from the \texttt{hosp} module of MIMIC-IV~v3.1~\cite{johnson2024mimiciv31, johnson2023mimicivscidata}, joined to the notes on the admission identifier. The first-listed code of the International Classification of Diseases (ICD) for the admission is taken as the primary diagnosis; 334 admissions carry an ICD-9 code and 221 an ICD-10 code. Both are grouped into the same 14 ICD chapters and the remaining chapters are pooled as Other. The fate of an error can then be reported by the kind of illness treated. All 555 admissions with eligible entries have a primary diagnosis.
    \item \textbf{The drug vocabulary:} Drug names come from the RxNorm full release of September 8,
    2026~\cite{nelson2011rxnorm}: 14{,}689 ingredient names, 3{,}670 precise-ingredient and 3{,}844
    multiple-ingredient names, 5{,}118 brand names linked to their ingredients by the tradename
    relation, and 101{,}558 synonym strings for these concepts. A labeled name resolves to a drug
    when it matches one of these strings exactly (6{,}355 of the 7{,}331 resolved entries), or by
    its first word (925), or as a single word inside a two-word ingredient name (\emph{glargine} in
    \emph{insulin glargine}; 51). A fourth rule, a two-word prefix of a longer ingredient name,
    applied to none of these entries. A name consisting only of a dose-form, route, or salt word
    (\emph{tablet}, \emph{oral}, \emph{sodium}) never resolves.
\end{itemize}

All MIMIC-IV data were accessed under PhysioNet credentialed data use
agreements~\cite{goldberger2000physionet,pollard2026physionet}. All extraction, including the
language models, ran on a local machine, and no note text was sent to an external service. Every
count in this paper was measured from the distributed files.

\subsection{Data preparation}
\label{sec:preparation}

\subsubsection{Perturbation target selection}

\begin{figure*}[t]
\centering
\includegraphics[width=\textwidth]{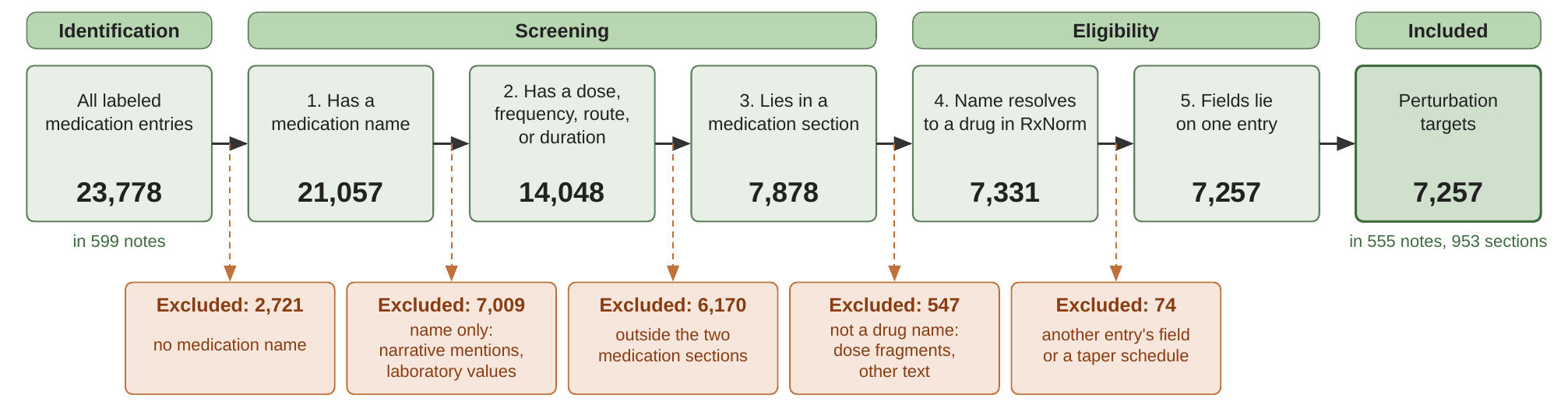}
\Description{A flow diagram from left to right under four stage labels: Identification,
Screening, Eligibility, and Included. The first box holds all 23,778 labeled medication entries
in 599 notes. Five boxes apply the inclusion criteria and give the entries kept: 21,057, 14,048,
7,878, 7,331, and 7,257. Under each arrow a box gives the number excluded and the reason: 2,721,
7,009, 6,170, 547, and 74. The last box holds the 7,257 perturbation targets in 555 notes.}
\caption{Selection flow for the perturbation targets. The five inclusion criteria are applied in
sequence, from left to right. Each box in the upper row gives the number of entries kept after
the criterion. The box under each arrow gives the number that the criterion excluded and the
reason. The criteria leave 7{,}257 perturbation targets in 555 notes and 953 sections.}
\label{fig:selection}
\end{figure*}

The study uses the 599 labeled notes; their median length is 9{,}936 characters. The character
offsets of the labels were verified against the note text before use. An error introduced at a
span that is not a medication would change nothing of clinical meaning, so five inclusion
criteria, applied in sequence, select the perturbation targets (Fig.~\ref{fig:selection}). An entry
must (1) have a medication name; (2) have at least one dose, frequency, route, or duration
field, because name-only entries are narrative mentions or laboratory values; (3) lie in a
Discharge Medications or Medications on Admission section, because these are the medication
lists the configurations read; (4) have a name that resolves to a drug in RxNorm, which removes
dose fragments and other non-drug text tagged as names without requiring that the drug was
ordered during the stay; and (5) have its dose, or its frequency when it has no dose, on the same
entry of the note as its name. Criterion~5 concerns 413 entries in which this field lies on a
different line from the labeled name. In 184 of them the entry names the drug twice, and the label marks the mention that is not on the
line of the dose. For these the name span is moved to the mention on that line: the numbered
line for 109, the \emph{RX} line for 59, and another line for 16. A further 155 are wrapped
entries, kept by treating an entry as a block from its numbered line to the next numbered or
blank line. The remaining 74 attach another entry's field or a taper schedule and are excluded.
The criteria leave 7{,}257 eligible entries in 555 notes from 554 patients; one patient
contributes two summaries, treated as independent. \mshi{One of the eligible entries receives no perturbation (Section~\ref{sec:injection}), which leaves 7{,}256 perturbed entries.} The eligible entries lie in 953 sections with a median length of 628 characters.

The labels of the eligible entries were audited by hand. One annotator, an author of this study,
judged the four labeled fields of 200 eligible entries drawn at random. A label is wrong when it
does not mark the value that the note gives for the field, and also when the note gives a value
and the label is missing. The name label was correct in all 200 entries, the dose label in 199,
the frequency label in 197 of 199 (one could not be decided), and the route label in 191
(95.5\%; 95\% Wilson interval~\cite{wilson1927probable} 91.7\% to 97.6\%). Nine of the 200
entries had at least one wrong label.

\subsubsection{Perturbations per class}
\label{sec:injection}

An entry can receive a perturbation only if it carries the field that the class edits.
Table~\ref{tab:candidates} reports, per class, how many eligible entries carry the field, how
many were perturbed, and why the rest were not.

\begin{table}[!b]
\caption{Perturbation targets per class. From left to right, each row shows the perturbation
class, the number of target fields (the field that the class edits, one in each eligible entry
that carries it), the number of fields perturbed, and the number not perturbed, with the reason.}
\label{tab:candidates}
\centering
\footnotesize
\setlength{\tabcolsep}{4pt}
\begin{tabular}{lrrl}
\toprule
Perturbation classes & \# of target fields & \# of perturbed fields & Not perturbed fields\\
\midrule
Drug-name substitution & 7{,}257 & 6{,}413 & 844 no confusable name \\
Tenfold dose       & 6{,}928 & 6{,}714 & 214 no number \\
Dose deletion      & 6{,}928 & 6{,}928 & 0 \\
Daily to QID       & 6{,}974 & 3{,}035 & 3{,}937 other frequency, 2 outside section \\
qhs to q1h         & 6{,}974 &   232   & 6{,}741 other frequency, 1 outside section \\
Frequency deletion & 6{,}974 & 6{,}968 & 6 outside section \\
Route deletion     & 6{,}281 & 6{,}269 & 12 outside section \\
\bottomrule
\end{tabular}
\end{table}

For the drug-name substitution, the ISMP partner supplied the substitute for 2{,}576 entries, the
composite score at 0.70 or above for 1{,}434, and the score at 0.55 to 0.69 for 2{,}403; 844
entries have no candidate at any tier. The candidates are the 1{,}226 RxNorm concepts in clinical
use here: those occurring in the labeled lists, in the cohort's orders, or on the ISMP list. For
the tenfold dose, entries with no dose span (329) or with no number in the dose span (214) are not
eligible. The three deletion classes reach 6{,}928, 6{,}968, and 6{,}269 entries. For 6 entries
the labeled frequency, and for 12 the labeled route, lies outside the section that contains the
drug name; these entries are not perturbed. One eligible entry carries only a duration and receives
no perturbation. The first frequency class reaches 3{,}035 entries; only 26 of them write the
abbreviation \emph{qd}, the string the documented misreading starts from, too few to report
separately. The second reaches 232 entries; the perturbed value is \emph{q1h}, because the corpus
writes every hour that way. Other frequency misreadings on the ISMP list have too few written
triggers in this corpus to use: \emph{qod} (every other day) appears in 16 eligible entries,
\emph{tiw} (three times a week) and \emph{q1d} (once daily) in none. Route confusion (SC read as
SL) is documented but not applied: only 77 eligible entries write the route as SC or SQ (both subcutaneous), too few to estimate a rate.

\subsection{\mshi{Extraction systems}}
\label{sec:pipelines}
An extraction system converts a medication section into structured fields: for each medication,
the drug name, dose, route, and frequency. The systems in this study belong to three categories.
The categories differ in how they find the fields and in how they assign the fields to a drug.
\mshi{Section~\ref{sec:configurations}} gives the versions and settings of the eight configurations.

\subsubsection{Dictionary-based}

cTAKES~\cite{savova2010ctakes} splits the text into sentences and
words. It compares single words and sequences of words with the terms
of a clinical dictionary that includes drug names and their synonyms.
A drug term that matches is identified as a medication and linked to
its code in the terminology. A name with no matching term in the
dictionary is not identified. cTAKES has a separate component for
drug attributes~\cite{savova2012drugner}, which this study runs
together with the lookup. The component searches the sentence that
contains the medication for a dose, a route, and a frequency. cTAKES
ends a sentence at every line break~\cite{ctakes2017core}, so the
sentence never extends beyond one line of the note. The component
recognizes each value by a fixed pattern, such as a number followed
by a unit for the dose, or a term from a list for the frequency. The
patterns are part of cTAKES and are not learned from data.

The dictionary lookup returns every term it finds, so one written name
can give several overlapping mentions: \emph{docusate sodium},
\emph{docusate}, and \emph{sodium}. The attribute component attaches the
dose, route, and frequency to only one of them, most often the last word.
This study combines overlapping mentions into one medication. The name
and its code are taken from the longest mention. Each attribute is taken
from the longest mention when it carries one, and otherwise from the
other mentions, starting with the one that begins last.

\subsubsection{Supervised tagger}
Med7~\cite{kormilitzin2021med7} is a named-entity recognition (NER)
model: a model trained on annotated text to mark spans of text and assign
a label to each. It was pretrained on more than two million MIMIC-III
documents. It was then trained on 1{,}212 documents: 909 MIMIC-III
discharge summaries annotated by hand and 303 documents labeled
automatically by pattern matching~\cite{kormilitzin2021med7}. Its seven
labels are drug, strength, dosage, form, route, frequency, and duration.
The strength label (\emph{500~mg}) is mapped to the dose field of this
study. The dosage label (\emph{one tablet}) is the quantity taken per
administration and has no counterpart among the four fields. The output
is a list of labeled spans with their positions in the text. Med7-trf has
the same labels and the RoBERTa-base-PM-M3
encoder~\cite{lewis2020pretrained}, which represents each word using the
other words in a window of the text. The encoder was pretrained on text
that includes MIMIC-III physician notes~\cite{lewis2020pretrained}. The
training corpus of Med7-trf is not described: its model card states only
that the model was retrained, and the configuration distributed with the
model refers to a MIMIC text file.

Med7 does not determine which spans belong to which drug. This study adds
a grouping rule after Med7. Each span that is not a drug is assigned to
the nearest drug before it in the same unit. A span with no drug before
it is dropped. For the route and the frequency, the first span in text
order is kept for a drug. For the strength, every span is kept, because
one entry can state two doses; \mshi{Section~\ref{sec:configurations}} gives the
rule that selects one. The unit is a single line in one configuration and
the whole medication entry in another.

\subsubsection{Language model}
Llama~\cite{llama3herd2024} and Qwen~\cite{qwen2025qwen25} receive an
instruction, the prompt, and the whole section. Each model generates its
reply one token at a time; a token may be a word, part of a word, or
punctuation. Each new token is predicted from the prompt, the section,
and the tokens already generated. The reply is parsed into the structured
fields. Unlike Med7, these models generate the field values and do not
mark spans in the section. The prompt of this study asks for field values
only, so the reply contains no positions and a returned value has no
explicit link to a place in the section.

\subsection{Extraction configurations}
\label{sec:configurations}

Four extraction systems of three categories are used in eight configurations
(Table~\ref{tab:configs}), to test whether propagation depends on how a system extracts the
fields. Med7 is run in two groupings, by line and by entry block, to test whether reading the
\emph{RX} line matters. In the block grouping, a second mention of the same drug within the
block, as on the \emph{RX} line, is treated as the same medication, and its spans are added to
those of the first mention. Med7 is also run with the transformer model that its authors
distribute (Med7-trf), grouped by line, to test whether the encoder of the tagger matters.
Llama~3.1 is run under three prompts and Qwen2.5 under the zero-shot prompt, to test the prompt
and the model family. Each configuration keeps the same settings for every section. The prompts
are given in Table~\ref{tab:prompts}. None mentions errors or asks the model to check anything.

\begin{table}[!b]
\caption{Extraction configurations. Each row is one of the eight configurations. From left to
right: the name used in the tables and figures, the extraction system with its category, and the
setting that distinguishes the configuration from the others of the same system.}
\label{tab:configs}
\centering
\footnotesize
\setlength{\tabcolsep}{4pt}
\begin{tabular}{@{}lll@{}}
\toprule
Configuration & System & Setting \\
\midrule
cTAKES    & cTAKES~\cite{savova2010ctakes}, dictionary-based & with drug attributes \\
Med7      & Med7~\cite{kormilitzin2021med7}, supervised tagger & grouped by line \\
Med7-B    & Med7                                       & grouped by entry block \\
Med7-trf  & Med7, transformer~\cite{lewis2020pretrained} & grouped by line \\
Llama ZS  & Llama~3.1 8B~\cite{llama3herd2024}, language model & zero-shot prompt \\
Llama 1S  & Llama~3.1 8B                               & one-shot prompt \\
Llama CoT & Llama~3.1 8B                               & chain-of-thought prompt \\
Qwen ZS   & Qwen2.5 7B~\cite{qwen2025qwen25}, language model & zero-shot prompt \\
\bottomrule
\end{tabular}
\end{table}

Two scoring rules are specific to one system. For cTAKES the drug name is compared together with
its RxNorm code, because a substitution that reaches the structured output carries a code as
well as a name; the pair counts as masked only when both the name and the code equal the
reference. For the Med7 configurations, when several strength spans are assigned to one drug,
the dose is taken from the span that overlaps the labeled dose position, and from the first span
when none overlaps. One exclusion is specific to the language models: a reply that is not valid
JSON after one retry is excluded and reported.

\subsubsection{Software and hardware}

Table~\ref{tab:software} lists the versions. Every run took place on one
Linux machine. cTAKES and Med7 ran on its CPU; the language models ran on
its NVIDIA RTX~6000 Ada GPU, served at 4-bit quantization
(\texttt{Q4\_K\_M}) with temperature~0, seed~42, a 4{,}096-token context
window, larger than the longest section, and a reply limit of 1{,}500
tokens. Every language-model configuration processed eight sections concurrently.

The three Med7 configurations returned identical outputs for all 37{,}512 inputs in a second
full run. cTAKES returned the same drug names, codes, and routes, but the dose or frequency it
attached to a drug differed in 1.7\% of the outputs when the input files were grouped
differently. No fate changed among its 21{,}315 scored pairs, and its inter-entry interference
rates changed by at most 0.1 percentage points.

For the language models, temperature~0 and a fixed seed do not make concurrent inference
reproducible. To measure the run-to-run variation, each language-model configuration extracted
every unperturbed section ten times with identical settings. Section~\ref{sec:results} reports
how often these runs agree.

\subsubsection{Prompts}

The three prompt conditions share one system instruction and differ in what follows it
(Table~\ref{tab:prompts}). The one-shot condition inserts one worked example as a user and
assistant turn before the section. The chain-of-thought condition appends a directive to the
system instruction: the model justifies each field by quoting the entry before it gives the JSON
array. The directive follows zero-shot chain-of-thought
prompting~\cite{kojima2022zeroshot} with the evidence-grounding step of
PromptNER~\cite{ashok2023promptner}. A reply that is not valid JSON is requested once more with
the line ``Your previous reply was not valid JSON. Reply with a JSON array only.''; a second
failure is counted as unparseable. Under the chain-of-thought condition the reasoning can use up
the reply limit of 1{,}500 tokens before the array is written. This happened in the first reply
for 13{,}188 of the 37{,}512 inputs (35.2\%), and never in the second. For these inputs the output
of this condition comes from the second request, which asks for the array only. In the other
three language-model configurations the limit was reached in five replies in total. The prompt's
key \emph{strength} is the field this study calls the dose.

\begin{table}[!t]
\caption{Prompt conditions. The first row quotes the system instruction shared by all three
conditions; the zero-shot condition uses it alone. The second row gives the example that the
one-shot condition adds. The third row quotes the directive that the chain-of-thought condition
appends to the system instruction.}
\label{tab:prompts}
\centering
\footnotesize
\begin{tabular}{@{}p{0.17\columnwidth}p{0.77\columnwidth}@{}}
\toprule
Condition & Content \\
\midrule
System (all) & You extract medications from a section of a hospital
discharge summary. Return ONLY a JSON array, no prose, no markdown. Each
element is an object with keys ``name'', ``strength'', ``route'',
``frequency''. Copy each value EXACTLY as it appears in the text, character
for character; do not correct, expand, abbreviate or normalize anything. Use
null for a value that is not stated. Include every medication entry in the
order it appears. If the text has no medications, return []. \\
\addlinespace
One-shot & Example user turn: ``Discharge Medications: 1. Lisinopril 10 mg
Tablet Sig: One (1) Tablet PO DAILY (Daily). 2. Atorvastatin 40 mg PO QHS'';
example assistant turn: the corresponding two-element JSON array. \\
\addlinespace
Chain of thought & Appended to the system instruction: Think step by step.
For each medication entry, in the order it appears, quote the entry as
written, then justify each field by quoting the exact word or phrase in that
entry that supports it: for strength, the number and unit; for route, the
abbreviation or word; for frequency, the abbreviation or phrase. If a field
is not present in the entry, write `not stated' instead of a justification.
After justifying every entry, output the JSON array on the last line, and
nothing after it. \\
\bottomrule
\end{tabular}
\end{table}

\begin{table}[!t]
\caption{Software versions. Each row gives a component and the version used for all runs of this
study. For the two language models the row gives the Ollama model tag, its identifier, the file
format, and the quantization.}
\label{tab:software}
\centering
\footnotesize
\begin{tabular}{@{}p{0.30\columnwidth}p{0.65\columnwidth}@{}}
\toprule
Component & Version / Ollama model \\
\midrule
Apache cTAKES & 4.0.0.1, Unified Medical Language System (UMLS) 2016AB (SNOMED~CT, RxNorm) \\
Med7 & 3.4.2.1 on spaCy 3.4.4 \\
Med7-trf & \texttt{en\_core\_med7\_trf} 1.1.0, RoBERTa-base-PM-M3 encoder, spaCy 3.8.18, spacy-transformers 1.4.0 \\
Ollama & 0.34.2 \\
Llama 3.1 8B Instruct & \texttt{llama3.1:latest} (ID 46e0c10c039e), GGUF, \texttt{Q4\_K\_M} \\
Qwen2.5 7B Instruct & \texttt{qwen2.5:7b} (ID 845dbda0ea48), GGUF, \texttt{Q4\_K\_M} \\
RxNorm & full release, September 8, 2026 \\
Python & 3.11 \\
\bottomrule
\end{tabular}
\end{table}

\subsection{\mshi{Measurement}}
\label{sec:measurement}
Each of the eight configurations is one extraction pipeline $\mathcal{T}(\cdot)$.

\subsubsection{Input}
Every configuration receives the medication section that contains the entry, from its header to
the next top-level header. The section is processed once unperturbed, which gives the reference
output for every entry in it, and once for each perturbation. The perturbed and reference
inputs therefore differ only at the edited span.

\subsubsection{Matched pairs}
Each perturbation is scored as a pair: the configuration's output on the perturbed section
against its output on the same section unperturbed. Both outputs are first mapped to one
representation, a list of medication records with the drug name, its character position, and the
dose, route, and frequency as extracted. The values are then normalized. Names are lowercased,
and salts and form words are removed. Doses are compared as a number and a unit. Frequencies and
routes are mapped to canonical forms (\emph{daily}, \emph{once a day}, and \emph{QD} all become
\emph{qd}; \emph{PO}, \emph{by mouth}, and \emph{oral} become \emph{po}). Dose and route words
copied together with a frequency (\emph{one tablet PO daily}) are removed before the mapping,
and the count-per-day form that cTAKES writes (\emph{2.0 day}) is mapped to the same forms
(\emph{bid}).

The edited entry is located in both outputs by its labeled drug-name position, so an entry whose
name has been substituted still matches its own record. The record chosen is the one whose name
position overlaps the labeled position the most. A language model returns no positions, so each
name it returns is located in the section text, in the order of the reply. The search for a name
starts after the previous located name. When the name does not occur there, the search restarts
at the start of the previous name. A returned name that does not occur in the text matches no
entry. Other entries in the section are aligned between the two outputs by their name position
as well, and an entry present in one output and absent from the other counts as changed. Correct
extraction means agreement with the label, not clinical correctness.

A pair is scored only when the reference output extracts the entry correctly: the drug name
matches the label, and for the six classes other than the drug-name substitution the field that the class edits matches the label too. Pairs that fail this condition are counted and reported but not
scored, so that masking is never confused with a configuration's inability to read the entry.
Entries whose edited field lies outside the entry block are excluded and reported as well.

\subsubsection{Fate assignment}

The fate is decided on the field that the class edits: the drug name for the drug-name
substitution, the dose for the tenfold dose and the deleted dose, the frequency for the two
frequency misreadings and the deleted frequency, and the route for the deleted route. For the
deletion classes the perturbed value is the absence of the field. For each scored deletion pair the
scorer also records whether a copy of the deleted value remains inside the entry's own block. A
dose is matched as text, and a frequency or a route by canonical form. Masking is then reported
separately for entries with and without such a copy. The canonical form of a frequency leaves
out text in parentheses, so \emph{DAILY (Daily)} becomes \emph{qd}. One rule is therefore added
for the two frequency misreadings: a pair is not masked when the frequency field of the perturbed
output holds the perturbed value in parentheses and the same field of the reference output does
not. Such a pair is scored like any other pair whose field changed; the rule applies to 63 pairs,
all of them daily to QID. Section~\ref{sec:configurations} gives the
scoring rules that are specific to one system.

\subsubsection{Interference}

Inter-entry interference is counted on the four extracted fields of the other entries in the
section: name, dose, route, and frequency. All entries a configuration returns are compared,
with or without a label. A language model can change other entries from one run to the next
without any edit. Each unperturbed section is therefore extracted ten times, and the
interference of a language model is compared with the differences between these repeat runs.
Intra-entry interference is measured on the three fields of the edited entry that the class does
not edit, compared after the same normalization as the fate. A pair whose entry is missing from
the perturbed output has no fields to compare and is counted separately.

\subsubsection{Difficulty strata}

Independently of the perturbation, each configuration either extracts an entry correctly from the unperturbed section or it does not. An entry is \emph{easy} when all configurations extract it correctly,
\emph{hard} when none does, and \emph{intermediate} when some do. Configurations are compared on
the easy entries, because there every configuration is scored on the same entries and a
difference between configurations can only come from how each handles the perturbation.
Pairwise differences in masking rate on the easy stratum are tested with McNemar's exact
test~\cite{mcnemar1947note}, with a Bonferroni correction for the number of pairwise tests per
class. With eight configurations each class has 28 pairwise tests, so a difference is reported
as significant \mshi{when $p$ is} below 0.0018 (0.05 divided by 28). Hard entries cannot be scored and are reported
as a count.

\subsubsection{Reported measures}

For each configuration and class, the study reports the number of pairs, the number excluded and
the reason, the number scored, and the count and share of each fate. The propagation \mshi{rate}
is the complement of the masking rate, and the direct and distorted shares partition it. For the
deletion classes the masking rate is also reported separately for pairs with and without a copy of
the deleted value in the entry's block. Intra-entry interference is reported as the
percentage of scored pairs in which another field of the edited entry changes\mshi{, among the pairs
whose edited entry is returned in the perturbed output}. Inter-entry
interference is reported as the percentage of scored pairs in which another entry changes. The
total error is reported at the level of the entry, $a + b$, and for the other entries of the
section, $c$. For the language models, the two interference rates, the entry-level total error,
and $c$ are compared with repeat runs of the unperturbed section. The 95\% intervals are
computed as follows:

\begin{itemize}
\item masking rate: the Wilson interval~\cite{wilson1927probable};
  because several pairs come from the same section, the rate is also
    bootstrapped over sections, with 1{,}000 replicates~\cite{efron1993bootstrap};
\item difference from repeat runs, in the two interference rates and
  in $c$: 1{,}000 bootstrap replicates that resample sections;
\item entry-level total error: 1{,}000 bootstrap replicates that
  resample pairs.
\end{itemize}

The measured stage ends at the configuration's structured output. The results are conditional on
correct extraction of the unperturbed entry: they describe how an error behaves among entries
the configuration already reads, not among all medication entries.

\begin{figure*}[!t]
\centering
\subfloat[Fate of the perturbation, as a share of all pairs of the class.]{%
  \includegraphics[width=\textwidth]{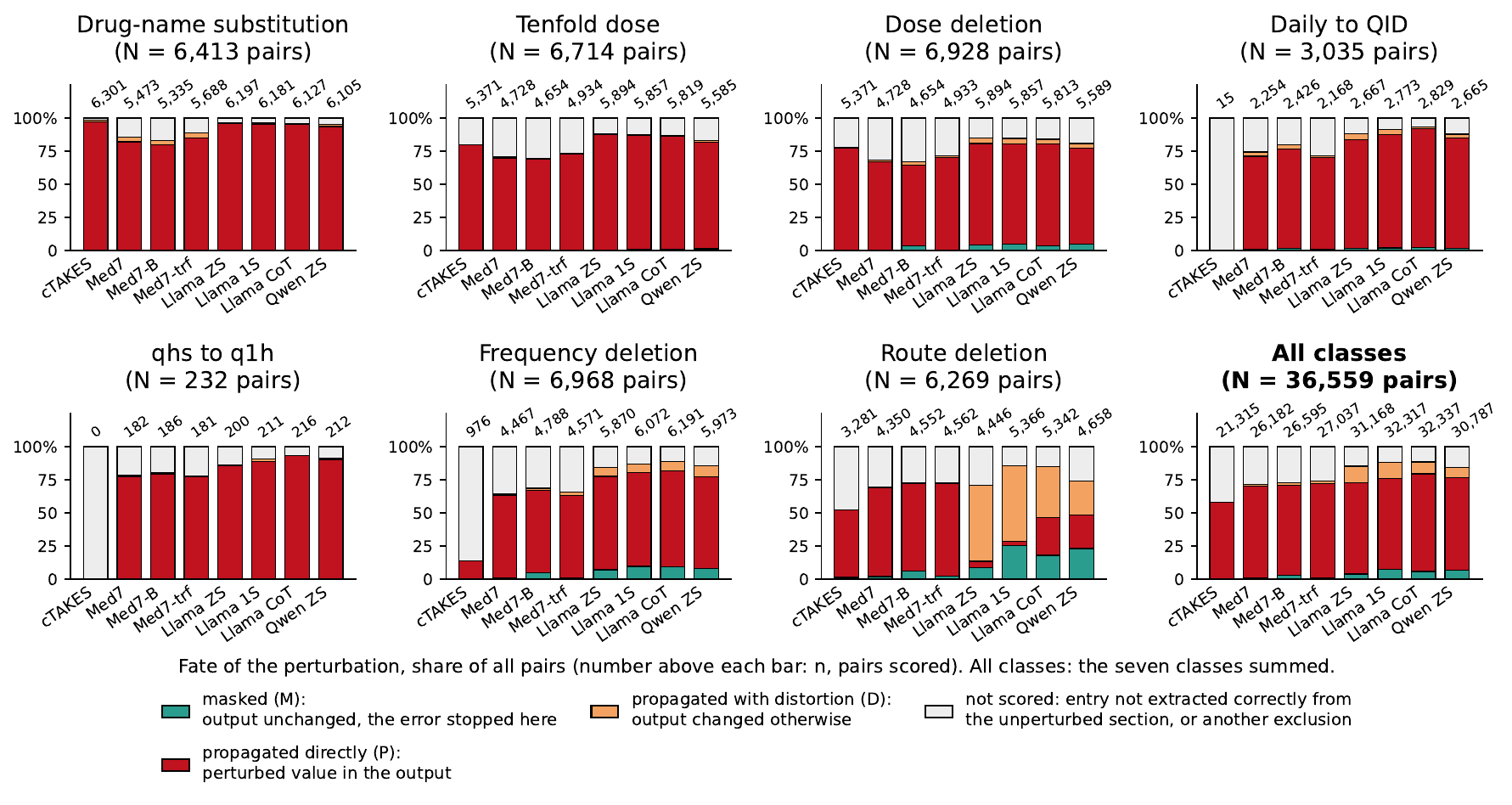}\label{fig:fates-a}}\\
\subfloat[Masked share of the scored pairs, with 95\% Wilson intervals.]{%
  \includegraphics[width=\textwidth]{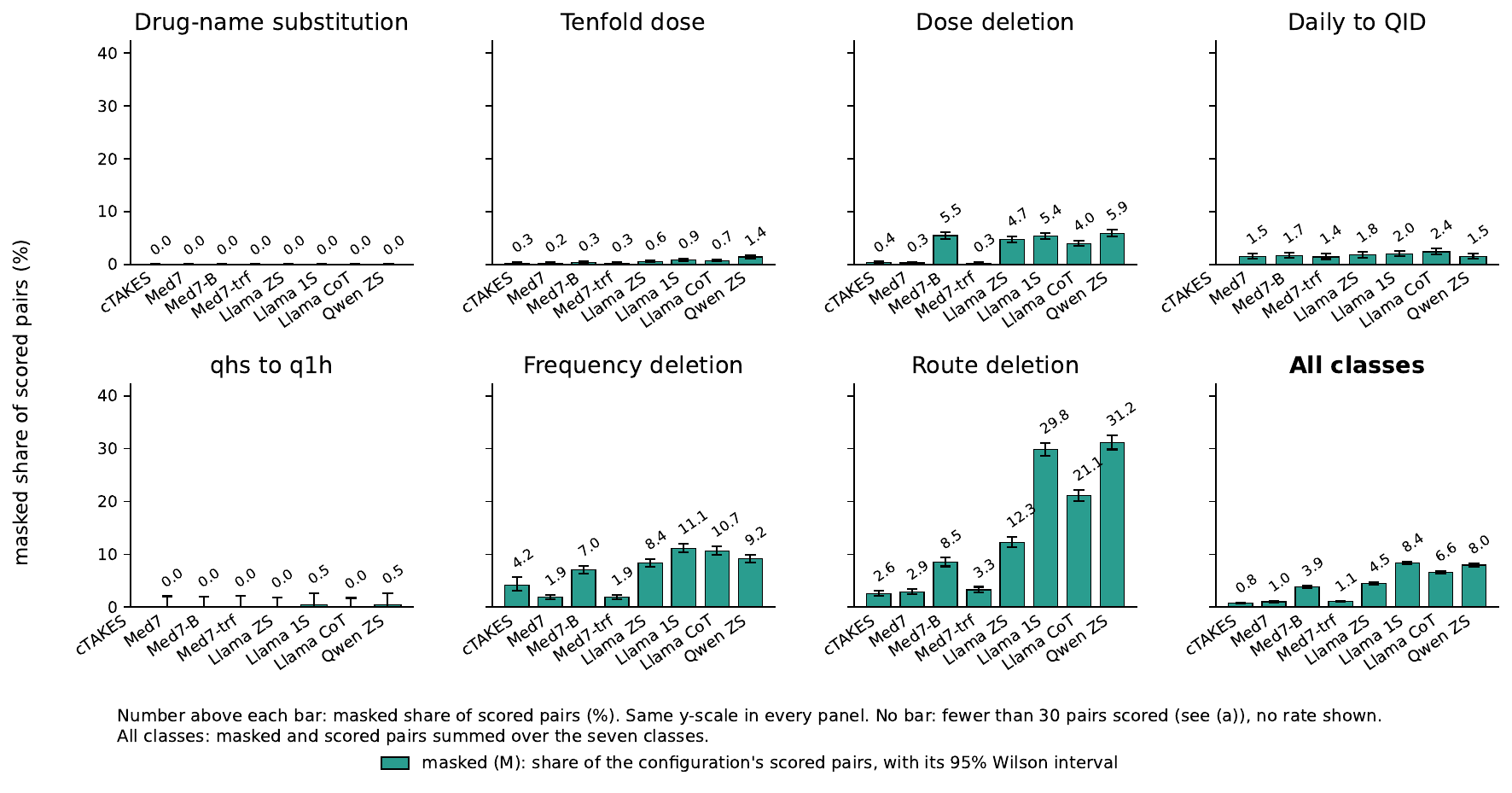}\label{fig:fates-b}}
\Description{Two rows of eight bar-chart panels, one panel per perturbation class and a last panel
for all classes, with one bar per configuration. In the upper row each bar shows the share of
pairs that were scored, divided into masked, propagated directly, and propagated with
distortion, and the unscored remainder. In the lower row each bar shows the masked share of the
scored pairs with its 95 percent interval.}
\caption{Fate of the perturbation by class and configuration. The last panel of each part sums the seven classes.}
\label{fig:fates}
\end{figure*}

\subsection{Run sizes and exclusions}
\label{sec:runs}

Every eligible entry was perturbed once per class for which it qualifies,
giving 36{,}559 perturbed copies of the 953 sections, one copy per
perturbation. Each of the 953 sections was also processed once unperturbed as the
reference. Every configuration processed all of them, which gives 36{,}559 pairs per
configuration. Table~\ref{tab:exclusions} reports, per
configuration, how many pairs were scored and how many were excluded for each
reason.

\begin{table}[!b]
\caption{Scored and excluded pairs per configuration. Each row is one configuration, and its five
counts add up to the 36{,}559 pairs, one for each perturbed field. The first four counts give the pairs excluded for each reason, which Section~\ref{sec:runs} defines; the last column gives the pairs scored.}
\label{tab:exclusions}
\centering
\footnotesize
\setlength{\tabcolsep}{3pt}
\begin{tabular}{lrrrrr}
\toprule
Configuration & Not extracted & Outside block & Unparseable & Empty output & Scored \\
\midrule
cTAKES    & 14{,}946 & 298 &   0 &  0 & 21{,}315 \\
Med7      & 10{,}079 & 298 &   0 &  0 & 26{,}182 \\
Med7-B    &  9{,}666 & 298 &   0 &  0 & 26{,}595 \\
Med7-trf  &  9{,}216 & 298 &   0 &  8 & 27{,}037 \\
Llama ZS  &  5{,}092 & 298 &   1 &  0 & 31{,}168 \\
Llama 1S  &  3{,}895 & 298 &  49 &  0 & 32{,}317 \\
Llama CoT &  3{,}715 & 295 & 167 & 45 & 32{,}337 \\
Qwen ZS   &  5{,}420 & 298 &  19 & 35 & 30{,}787 \\
\bottomrule
\end{tabular}
\end{table}

``Not extracted'' counts pairs whose reference output did not extract the entry correctly
(Section~\ref{sec:measurement}); these are reported, not scored. cTAKES has the most, because it
attaches a frequency to few entries in this note format, so its frequency classes are mostly
unscorable (15, 0, and 976 scored pairs). Fig.~\ref{fig:fates-a} shows the unscored share per
class. ``Outside block'' counts pairs whose edited field lies outside the entry block.
Criterion~5 (Section~\ref{sec:preparation}) tests one field of an entry, so a frequency or a route
can still lie outside the block. This holds for 216 entries, and their 298 pairs per configuration
are in the three frequency classes (163) and in the route deletion (135).
``Unparseable'' counts pairs with a language-model reply that was not valid JSON after one retry. They are
concentrated in few sections: 48 of the 49 pairs of Llama 1S and 18 of the 19 pairs of Qwen ZS
come from one section each, and 140 of the 167 pairs of Llama CoT from two of twelve sections.
``Empty output'' counts pairs whose perturbed output returned no entries while the reference
output returned at least two; this is an output failure of the configuration, not a fate. The
exclusions are applied in the order unparseable, empty output, outside block, not extracted, and
a pair is counted under the first that applies.

\mshi{The fates assigned to the scored pairs were checked} by hand on a sample. One annotator
judged 240 scored pairs, 80 drawn at random from each fate under the exact criterion, without
seeing the assigned fate. The sample was drawn from an earlier run of the eight configurations;
in the final runs, 230 of these 240 pairs have the same fate. The annotator agreed with the
fates of that earlier run on 218 pairs (90.8\%; 95\% Wilson interval 86.5\% to 93.9\%). The count
is the same under both criteria: of the 20 sampled pairs that the two criteria score differently, the annotator marked 5 as direct, 5 as distorted, and 10 as masked. Twenty of the 22 disagreements were in the deleted-frequency class, most of them a field that kept other words after the frequency was deleted. The token-level criterion (Section~\ref{sec:types}) was defined from these pairs.

\section{Results}
\label{sec:results}
\mshi{The results are reported in four parts: the fate of the perturbation (Section~\ref{sec:res-fate}), the comparison of the configurations (Section~\ref{sec:res-configs}), the changes outside the edited field (Section~\ref{sec:res-outside}), and the comparison of the drug-name similarity measures (Section~\ref{sec:similarity}).}

\subsection{\mshi{Fate of the perturbation}}
\label{sec:res-fate}

\subsubsection{Replaced values: drug name, dose, and frequency}

Fig.~\ref{fig:fates-a} shows, for each class and configuration, the
share of pairs that the configuration scored. Within the
scored share it shows the share masked, propagated directly, or
propagated with distortion. The unscored share is mainly the pairs whose entry the
configuration did not extract correctly from the unperturbed section;
Table~\ref{tab:exclusions} gives the other exclusions.
Over all classes it is largest for cTAKES (Section~\ref{sec:runs}).
The rates that follow are over scored pairs; Fig.~\ref{fig:fates-b}
shows them with their intervals. Four classes propagate through every
configuration that scores them. A substituted drug name is masked in 0.0\% of pairs in all
eight configurations, on 5{,}335 to 6{,}301 pairs each; cTAKES
assigns the substituted name its own RxNorm code. A tenfold dose is
masked in 0.2\% to 1.4\% of pairs. The two frequency misreadings
behave the same way. In the seven configurations other than cTAKES,
the masked share is 1.4\% to 2.4\% for \emph{daily} to \emph{qid} and
0.0\% to 0.5\% for \emph{qhs} to \emph{q1h}. cTAKES scores 15 and 0
pairs in these two classes and is not reported.

The last panel of Fig.~\ref{fig:fates-a} and the last panel of Fig.~\ref{fig:fates-b} sum the
seven classes. In the last panel of Fig.~\ref{fig:fates-a} the denominator is the 36{,}559 pairs
of the configuration. In the last panel of Fig.~\ref{fig:fates-b} it is the configuration's
scored pairs of all seven classes, so a class counts in proportion to its scored pairs. In that
panel, the masked share over all classes ranges from 0.8\% for cTAKES to 8.4\% for Llama 1S. One
medication entry is perturbed once for each class it is eligible for, so the pairs summed in the
two last panels are not independent, and the Wilson interval in the last panel of
Fig.~\ref{fig:fates-b} understates the uncertainty.

\begin{figure*}[!t]
\centering
\includegraphics[width=\textwidth]{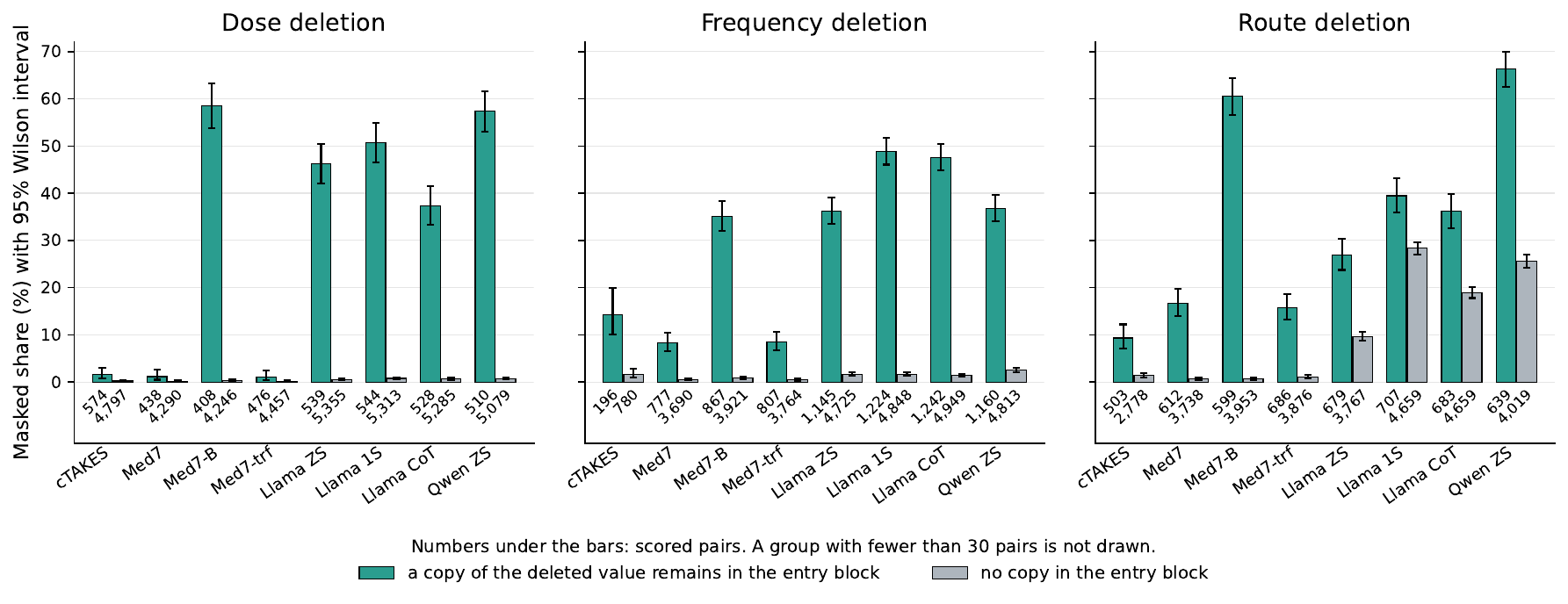}
\Description{Three bar-chart panels, one for each class in which a dose, a frequency, or a route
is deleted. For every configuration two bars show the masked share of scored pairs with 95
percent intervals: one for entries that still hold a copy of the deleted value in their block and
one for entries without a copy. The number of scored pairs is printed under each bar.}
\caption{Masked share of the three deletion classes, with 95\% Wilson
intervals. Pairs are split by whether a copy of the deleted value remains in
the entry's block: its numbered line and the \emph{RX} line beneath it.
Numbers under the bars are scored pairs.}
\label{fig:redundancy}
\end{figure*}

\subsubsection{Deleted values}

Over all scored pairs of a class, masking above 3\% occurs only in the deletion classes, and
Fig.~\ref{fig:redundancy} shows why. When the dose is deleted, the language models and Med7-B return it in 4.0\% to 5.9\% of pairs overall, while Med7, Med7-trf, and cTAKES return it in 0.3\% to 0.4\%. The section-level
bootstrap widens the intervals (Llama ZS 4.7\%, 3.9\% to 5.6\% over sections against
4.2\% to 5.3\% over pairs) without changing the ordering. An entry can hold a second copy of the
deleted value in its block: on the pharmacy \emph{RX} line, or on the numbered line itself,
where some entries state a value twice (\emph{DAILY (Daily)}). With such a copy, the masked
share is 37.3\% to 58.6\% for the language models and Med7-B, and it is below 1\% without one.
Med7, Med7-trf, and cTAKES mask under 2\% even with a copy, because they do not read the
\emph{RX} line as part of the entry. Within Med7, the grouping alone changes the result: Med7-B
masks 58.6\% of deleted doses with a copy, Med7 and Med7-trf 1.1\%. A deleted frequency follows
the same pattern: 35.1\% to 48.9\% masked with a copy by the language models and Med7-B, under
3\% without. A deleted route differs: the language models mask 9.7\% to 28.3\% of pairs even
without a copy, filling the field with a default such as \emph{PO}, while Med7, Med7-B,
Med7-trf, and cTAKES mask 0.6\% to 1.4\% without a copy. The groups with a copy are small: 408
to 574 dose pairs per configuration.

Table~\ref{tab:template} splits the masked share by note template. Of the 7{,}256 perturbed
entries, 730 are in the \emph{RX}-line template and 6{,}526 in the \emph{Sig:}
template. For a deleted dose, the language models and Med7-B mask 39.3\% to 62.8\% of
\emph{RX}-line entries and at most 1.0\% of \emph{Sig:} entries. Med7, Med7-trf, and cTAKES mask
at most 0.9\% in both, because they do not read the \emph{RX} line. A deleted frequency follows
the same pattern for the language models and Med7-B (28.6\% to 54.2\% against 2.7\% to 9.1\%).
\mshi{For the daily-to-QID misreading the masked share is also higher in the \emph{RX}-line
template than in the \emph{Sig:} template, 10.2\% to 16.3\% against 0.6\% to 1.5\% over the seven
scored configurations. Med7 and Med7-trf, which do not read the \emph{RX} line, lie within these
ranges.} A deleted route is the one case masked in the \emph{Sig:} template: the language models
mask 11.3\% to 29.2\% of those entries by supplying a default route, while Med7-B masks 77.2\% of
\emph{RX}-line entries and 2.4\% of \emph{Sig:} entries. For a deleted frequency, a deleted
route, and daily to QID, Med7, Med7-trf, and cTAKES mask 6.6\% to 13.0\% of \emph{RX}-line
entries. They do not take the value from another line. In 256 of their 278 masked pairs, the
value appears a second time on the line of the labeled name. When the copy is only on another
line of the entry, they mask at most 0.9\% of pairs.

\begin{table*}[!t]
\caption{Masked share (\%) by note template for the deletion classes and the daily-to-QID
misreading. Each row is one configuration. For each of the four classes, the two columns give the
masked share of the scored pairs whose entry is in the \emph{Sig:} template (no \emph{RX} line) and
in the \emph{RX}-line template (a numbered line followed by a pharmacy \emph{RX} line).
-- : fewer than 30 scored pairs.}
\label{tab:template}
\centering
\footnotesize
\setlength{\tabcolsep}{5pt}
\begin{tabular}{lrrrrrrrr}
\toprule
 Configuration & \multicolumn{2}{c}{Dose deletion} & \multicolumn{2}{c}{Frequency deletion} & \multicolumn{2}{c}{Route deletion} & \multicolumn{2}{c}{Daily to QID} \\
 & Sig & RX & Sig & RX & Sig & RX & Sig & RX \\
\midrule
cTAKES    &  0.3 &  0.9 & 3.5 &  8.3 &  2.0 &  6.6 & --  & --   \\
Med7      &  0.2 &  0.6 & 1.1 &  8.7 &  1.8 & 13.0 & 0.6 & 11.2 \\
Med7-B    &  0.6 & 57.8 & 2.7 & 54.2 &  2.4 & 77.2 & 0.9 & 12.9 \\
Med7-trf  &  0.3 &  0.4 & 0.9 &  9.0 &  2.3 & 11.5 & 0.6 & 10.2 \\
Llama ZS  &  0.7 & 48.8 & 6.4 & 30.8 & 11.3 & 20.9 & 0.7 & 16.3 \\
Llama 1S  &  1.0 & 52.7 & 8.8 & 37.1 & 29.2 & 36.1 & 1.1 & 15.3 \\
Llama CoT &  0.8 & 39.3 & 9.1 & 28.6 & 20.2 & 31.0 & 1.5 & 15.3 \\
Qwen ZS   &  0.9 & 62.8 & 6.1 & 44.8 & 26.4 & 78.0 & 0.8 & 12.5 \\
\bottomrule
\end{tabular}
\end{table*}

\subsubsection{Dependence on the admission diagnosis}

This analysis examines whether the illness treated affects how often a deleted dose is masked.
It uses the deleted dose because masking in this class depends almost entirely on whether the
entry repeats the dose. Table~\ref{tab:diagnosis} groups the pairs by primary-diagnosis chapter
and combines the four language-model configurations. Masking ranges from 2.1\% for Blood to
8.2\% for Endocrine. Two factors account for this variation. The first is how often the entry
repeats the dose \mshi{in its block} (5.4\% to 11.4\% of pairs). The second is how often the
language models mask the deletion when such a copy exists (19\% to 55\%). \mshi{In 13 of the 15
chapters the masked share is within one percentage point of the product of the two, computed from the unrounded shares; Endocrine (8.2\% against a product of 5.5\%) and Infectious (4.9\% against 3.5\%) are the exceptions. The
product is the share of pairs masked with a copy, so the remainder is masked without one; it
exceeds one percentage point only in these two chapters.} A
chapter can repeat the dose often and still mask little, as Blood does. Few
pairs in each chapter have a copy, so the rates among them are imprecise. Without a copy, masking
is rare in every chapter. The differences between chapters therefore reflect mainly how often the
note repeats the dose, not the diagnosis itself.

\begin{table}[!b]
\caption{Masking of a deleted dose by primary-diagnosis chapter, for the four language-model
configurations pooled. From left to right, each row shows the diagnosis chapter, the number of
admissions, the number of scored pairs, the share of pairs masked, the share of pairs whose entry repeats the dose, and the share masked among the pairs with such a copy. Rows are ordered by the
masked share, from lowest to highest.}
\label{tab:diagnosis}
\centering
\footnotesize
\setlength{\tabcolsep}{3.5pt}
\begin{tabular}{lrrrrr}
\toprule
Chapter & Adm. & Pairs & Masked (\%) & Copy (\%) & Masked with copy (\%)\\
\midrule
Blood            &  12 &   562 &  2.1 & 11.0 & 19.4 \\
Symptoms         &  38 & 1{,}887 &  2.6 &  5.4 & 42.6 \\
Mental           &  11 &   481 &  2.7 &  9.6 & 28.3 \\
Other            &  15 &   600 &  3.2 &  7.0 & 45.2 \\
Genitourinary    &  30 & 1{,}342 &  3.6 &  7.9 & 41.5 \\
Skin             &  14 &   712 &  3.7 &  9.4 & 38.8 \\
Digestive        &  80 & 2{,}525 &  4.6 &  9.0 & 48.0 \\
Injury           &  63 & 2{,}457 &  4.8 &  8.4 & 50.5 \\
Infectious       &  31 & 1{,}269 &  4.9 &  7.5 & 47.4 \\
Nervous          &  22 &   913 &  4.9 &  8.7 & 48.1 \\
Respiratory      &  36 & 1{,}693 &  5.1 &  9.0 & 47.1 \\
Musculoskeletal  &  30 & 1{,}299 &  5.1 &  9.2 & 45.8 \\
Neoplasm         &  53 & 1{,}832 &  5.4 & 10.0 & 51.4 \\
Circulatory      &  97 & 4{,}526 &  6.8 & 11.4 & 55.0 \\
Endocrine        &  23 & 1{,}055 &  8.2 & 11.3 & 48.7 \\
\bottomrule
\end{tabular}
\end{table}

\subsubsection{Propagation with distortion}

Distortion occurs in every configuration (Fig.~\ref{fig:fates-a}, amber segments). For the
drug-name substitution class the distorted share is 0.6\% to 4.4\%, and for the tenfold dose
0.2\% to 1.5\%. In both classes the entry is missing from the output in most distorted pairs
(883 of 995 and 234 of 303). The deletion classes separate the configurations. When the frequency
is deleted, cTAKES distorts 0.1\% of pairs, Med7, Med7-B, and Med7-trf 0.8\%, 2.7\%, and 3.3\%,
the Llama configurations 7.8\% to 8.1\%, and Qwen ZS 9.6\%. When the route is deleted, cTAKES and
the three Med7 configurations leave the field empty (distorted share at most 0.5\%). The language
models fill it with other text of the entry, most often its frequency. Their distorted share is
80.9\% for Llama ZS, 66.5\% for Llama 1S, 45.3\% for Llama CoT, and 34.8\% for Qwen ZS.

Table~\ref{tab:distortionkinds} gives the type of each distorted pair, by class and
configuration, following Definition~\ref{def:kinds}. For a drug-name substitution, every
distortion of cTAKES and almost every one of the three Med7 configurations is the entry missing
from the output. In the language models, 15 to 76 pairs are a missing entry, and in the other 20
to 30 the name field holds the value of another field or text that occurs nowhere in the section.
When the route is deleted, 9{,}837 of the 11{,}205 language-model distortions hold the entry's
frequency in the route field, 71 hold its name or its dose, and 1{,}122 hold other text of the entry's \mshi{block}. When the frequency is deleted, the field holds other text of the entry's \mshi{block} in 215 to 417
language-model pairs and in 34 to 125 pairs of the three Med7 configurations. The language
models also return 35 to 177 values that occur nowhere in the section. When the dose is
deleted, the field holds other text of the entry's \mshi{block} in 142 to 221 language-model
pairs and in 120 Med7-B pairs. For the two frequency misreadings, most
distorted pairs of the configurations together are a missing entry (430 of 606 and 13 of 13).

Table~\ref{tab:criteria} compares the two criteria of Section~\ref{sec:types}. Under the
\emph{exact} criterion an error is propagated directly only when the field equals the perturbed
value. Under the \emph{token-level} criterion it is also propagated directly when the field holds
the perturbed value and no new word. Masking is the same under both, because the first case of
Eq.~\eqref{eq:fate} does not depend on the criterion. The criteria differ in 4{,}739 of the
227{,}738 scored pairs (2.1\%). Each of these pairs is distorted under the \emph{exact} criterion
and direct under the \emph{token-level} criterion. A deleted frequency accounts for 4{,}062 of
them: the field keeps words that it already held in the reference output. In the seven
configurations other than cTAKES, its distorted share is 7.6\% to 26.1\% under the \emph{exact}
criterion and 0.8\% to 9.6\% under the \emph{token-level} criterion. For the two dose classes the
criteria give the same fate for every pair. In the other four classes the distorted share
differs by at most 1.0 percentage points.

\begin{table*}[!t]
\caption{Type of distortion by class and configuration (number of distorted pairs). Each row is
one configuration within a class. Missing: the field or the entry is not returned. Other field:
the value of another field of the same entry. Entry text: other text of the entry's block. Other
entry: text outside the entry's block. Produced: text that occurs nowhere in the section. The last
column is the total number of distorted pairs. A configuration with no distorted pair in a class
has no row.}
\label{tab:distortionkinds}
\centering
\footnotesize
\setlength{\tabcolsep}{5pt}
\begin{tabular}{llrrrrrr}
\toprule
Class & Configuration & Missing & Other field & Entry text & Other entry & Produced & Distorted \\
\midrule
Drug-name substitution & cTAKES & 79 & 0 & 0 & 0 & 0 & 79 \\
 & Med7 & 214 & 0 & 0 & 0 & 3 & 217 \\
 & Med7-B & 209 & 0 & 0 & 0 & 2 & 211 \\
 & Med7-trf & 241 & 1 & 0 & 0 & 8 & 250 \\
 & Llama ZS & 19 & 13 & 0 & 0 & 12 & 44 \\
 & Llama 1S & 30 & 11 & 0 & 0 & 12 & 53 \\
 & Llama CoT & 15 & 9 & 0 & 0 & 11 & 35 \\
 & Qwen ZS & 76 & 18 & 0 & 0 & 12 & 106 \\
\addlinespace
Tenfold dose & cTAKES & 12 & 0 & 0 & 0 & 0 & 12 \\
 & Med7 & 34 & 0 & 2 & 1 & 6 & 43 \\
 & Med7-B & 26 & 0 & 3 & 1 & 3 & 33 \\
 & Med7-trf & 47 & 0 & 1 & 4 & 2 & 54 \\
 & Llama ZS & 20 & 1 & 5 & 4 & 1 & 31 \\
 & Llama 1S & 15 & 0 & 9 & 2 & 1 & 27 \\
 & Llama CoT & 12 & 0 & 2 & 1 & 3 & 18 \\
 & Qwen ZS & 68 & 0 & 15 & 1 & 1 & 85 \\
\addlinespace
Dose deletion & cTAKES & 3 & 0 & 11 & 5 & 0 & 19 \\
 & Med7 & 76 & 0 & 16 & 0 & 0 & 92 \\
 & Med7-B & 76 & 0 & 120 & 0 & 0 & 196 \\
 & Med7-trf & 35 & 0 & 26 & 0 & 0 & 61 \\
 & Llama ZS & 24 & 71 & 196 & 7 & 0 & 298 \\
 & Llama 1S & 21 & 34 & 221 & 11 & 0 & 287 \\
 & Llama CoT & 22 & 25 & 200 & 11 & 0 & 258 \\
 & Qwen ZS & 60 & 40 & 142 & 12 & 1 & 255 \\
\addlinespace
Daily to QID & Med7 & 88 & 0 & 4 & 0 & 1 & 93 \\
 & Med7-B & 92 & 0 & 12 & 0 & 3 & 107 \\
 & Med7-trf & 32 & 0 & 3 & 0 & 1 & 36 \\
 & Llama ZS & 85 & 1 & 17 & 0 & 30 & 133 \\
 & Llama 1S & 77 & 1 & 17 & 0 & 27 & 122 \\
 & Llama CoT & 14 & 0 & 8 & 0 & 9 & 31 \\
 & Qwen ZS & 42 & 1 & 13 & 0 & 28 & 84 \\
\addlinespace
qhs to q1h & Med7 & 2 & 0 & 0 & 0 & 0 & 2 \\
 & Med7-B & 2 & 0 & 0 & 0 & 0 & 2 \\
 & Med7-trf & 1 & 0 & 0 & 0 & 0 & 1 \\
 & Llama ZS & 1 & 0 & 0 & 0 & 0 & 1 \\
 & Llama 1S & 5 & 0 & 0 & 0 & 0 & 5 \\
 & Qwen ZS & 2 & 0 & 0 & 0 & 0 & 2 \\
\addlinespace
Frequency deletion & cTAKES & 0 & 0 & 0 & 1 & 0 & 1 \\
 & Med7 & 3 & 0 & 34 & 0 & 0 & 37 \\
 & Med7-B & 3 & 0 & 125 & 0 & 0 & 128 \\
 & Med7-trf & 47 & 3 & 57 & 0 & 45 & 152 \\
 & Llama ZS & 10 & 19 & 314 & 11 & 109 & 463 \\
 & Llama 1S & 16 & 33 & 298 & 26 & 102 & 475 \\
 & Llama CoT & 9 & 15 & 417 & 25 & 35 & 501 \\
 & Qwen ZS & 42 & 44 & 215 & 93 & 177 & 571 \\
\addlinespace
Route deletion & Med7 & 0 & 0 & 2 & 0 & 1 & 3 \\
 & Med7-B & 0 & 0 & 8 & 0 & 0 & 8 \\
 & Med7-trf & 11 & 0 & 13 & 0 & 0 & 24 \\
 & Llama ZS & 8 & 3{,}377 & 190 & 12 & 9 & 3{,}596 \\
 & Llama 1S & 12 & 3{,}214 & 307 & 26 & 8 & 3{,}567 \\
 & Llama CoT & 10 & 2{,}052 & 336 & 17 & 5 & 2{,}420 \\
 & Qwen ZS & 17 & 1{,}265 & 289 & 43 & 8 & 1{,}622 \\
\bottomrule
\end{tabular}
\end{table*}

\begin{table}[t]
\caption{Distorted pairs under the two criteria, eight configurations pooled. From left to right,
each row shows the perturbation class, the number of scored pairs, the share of them propagated
with distortion under the exact criterion and under the token-level criterion, and the number of
pairs to which the two criteria assign different fates. The last row covers all seven classes.}
\label{tab:criteria}
\centering
\footnotesize
\setlength{\tabcolsep}{4pt}
\begin{tabular}{lrrrr}
\toprule
Class & Scored pairs & Distorted, \emph{exact} (\%) & Distorted, \emph{token-level} (\%) & Pairs that differ \\
\midrule
Drug-name substitution & 47{,}407 &  2.6 &  2.1 &   258 \\
Tenfold dose      & 42{,}842 &  0.7 &  0.7 &     0 \\
Dose deletion     & 42{,}839 &  3.4 &  3.4 &     0 \\
Daily to QID      & 17{,}797 &  4.0 &  3.4 &   103 \\
qhs to q1h        &  1{,}388 &  1.9 &  0.9 &    13 \\
Frequency deletion & 38{,}908 & 16.4 &  6.0 & 4{,}062 \\
Route deletion    & 36{,}557 & 31.6 & 30.7 &   303 \\
\midrule
All classes       & 227{,}738 &  9.5 &  7.4 & 4{,}739 \\
\bottomrule
\end{tabular}
\end{table}

\subsection{\mshi{Comparison across configurations}}
\label{sec:res-configs}

\subsubsection{Prompt and model family}

On five classes the three prompts give masking rates within 1.5 percentage points of one another
(deleted dose 4.0\% to 5.4\%, tenfold dose 0.6\% to 0.9\%). For a deleted frequency, Llama ZS
masks 8.4\%, Llama 1S 11.1\%, and Llama CoT 10.7\%. For a deleted route they differ most:
Llama 1S masks the most (29.8\%), Llama CoT less (21.1\%), and Llama ZS the least (12.3\%).
Of the three prompts, Llama CoT also gives the smallest distorted share for a deleted route. The masking rates of Qwen ZS lie within 0.6 percentage points of the Llama range on six classes. For a deleted route
it masks more than any Llama configuration (31.2\%).

\subsubsection{Comparison on the easy stratum}

The easy stratum (Section~\ref{sec:measurement}) holds 4{,}643 entries for the drug-name
substitution, 3{,}706 for the tenfold dose and for a deleted dose, 594 for a deleted frequency,
and 2{,}547 for a deleted route. The two frequency misreadings have 7 and 0 easy entries, because
cTAKES attaches a frequency to few entries in this note format. The hard stratum holds 0 entries
for the drug-name substitution, 537 for the tenfold dose, 751 for a deleted dose, 281 for a
deleted frequency, 344 for a deleted route, and 66 and 4 for the two frequency misreadings.

On the easy stratum the masking rates for the drug-name substitution, the tenfold dose, and a
deleted dose are close to those on all scored pairs. For a deleted frequency and a deleted route
the language models mask less on the easy stratum than on all scored pairs: 4.2\% to 6.7\%
against 8.4\% to 11.1\%, and 7.5\% to 26.1\% against 12.3\% to 31.2\%. A deleted frequency is
masked in 1.9\% by Med7 and by Med7-trf, 2.9\% by cTAKES, and 7.7\% by Med7-B. A deleted route is
masked least by Med7 (0.9\%), Med7-trf (1.1\%), and cTAKES (1.4\%), and most by Qwen ZS (26.1\%);
Med7-B masks 9.3\%.

Differences are tested at the corrected threshold of 0.0018. For a deleted dose, McNemar's exact
test separates two groups: each language-model configuration and Med7-B differs from Med7,
Med7-trf, and cTAKES ($p < 10^{-41}$), and these three do not differ from one another. Within
the first group, only Llama CoT differs from the others ($p < 0.001$). For a
deleted frequency, Med7-B differs from Med7, Med7-trf, cTAKES, and Llama ZS
($p < 0.001$), and not from Llama 1S, Llama CoT, or Qwen ZS. For a deleted route,
\mshi{the order of the configurations by masking rate is} Qwen ZS, Llama 1S, Med7-B, Llama CoT, Llama ZS, cTAKES, Med7-trf, and Med7. Three of the seven adjacent comparisons differ ($p < 10^{-10}$): Qwen ZS against Llama 1S, Llama 1S against Med7-B, and Llama ZS against cTAKES.

\subsection{\mshi{Changes outside the edited field}}
\label{sec:res-outside}

\subsubsection{Run-to-run variation}

\mshi{Under concurrent inference, the} language models do not return the same output for the same section in every run. On 50
unperturbed sections, the reply of Llama ZS with eight sections processed concurrently
was identical to its reply in a sequential run for 42 sections. Two runs with eight concurrent
sections gave identical replies for 44. Over the ten runs of each unperturbed section, the
parsed output was identical in all ten for 67.9\% of sections with Llama ZS, 66.5\% with Llama
1S, 70.1\% with Llama CoT, and 53.9\% with Qwen ZS. Two runs gave identical parsed output for
81.8\% to 88.8\% of sections with the Llama configurations and for 73.4\% to 80.8\% with Qwen ZS.
Sections with an unparseable output in any run are excluded (at most six of 953 per
configuration).

The fates do not depend on which run serves as the reference. Scored against each of the ten
runs in turn, the masking rate of every language-model configuration and class ranges over at
most 0.3 percentage points. The share propagated with distortion ranges over at most 1.0
points, except for \emph{qhs} to \emph{q1h}. That class has 199 to 217 scored pairs, depending
on the reference run, and its distorted share ranges over at most 1.4 points.

\subsubsection{Interference}

Table~\ref{tab:interference} reports inter-entry interference as the percentage of scored pairs
in which at least one other entry changes, over all returned entries
(Section~\ref{sec:measurement}). cTAKES, Med7, and Med7-B change another entry in at most 4.6\%
of pairs, except Med7-B for the drug-name substitution (7.2\%). Med7-trf changes another entry in 7.5\% to
45.6\% of pairs. Med7-trf is deterministic: two full runs gave identical output for all
37{,}512 inputs. Its interference is therefore caused by the edit. Med7-trf and Med7 use the
same grouping rule and differ in the model; Section~\ref{sec:discussion} discusses this
difference.

The language models have two rows in the table. The upper row compares the output for the
perturbed section with each of the ten runs of the unperturbed section. The row ``repeat runs''
compares those ten runs with each other for the same pairs. Repeat runs alone change another
entry in 11.4\% to 18.0\% of comparisons for the Llama configurations and in 29.0\% to 30.0\%
for Qwen ZS. Each such change is an error in at least one of the runs, so a language model
introduces errors into other entries whether or not the section contains a perturbation.
The edit adds to these rates. Here and for the total error, the additions are computed
from the unrounded values, so one can differ in its last digit from the difference of two table cells.
For the drug-name substitution and the three deletion classes it adds 9.4 to
20.9 percentage points for the Llama configurations and 14.9 to 21.8 for Qwen ZS. For daily to QID
it adds 6.3 to 10.1 and 9.5 percentage points, and for the tenfold dose 3.3 to 5.3 and 6.2. In
these six classes the 95\% bootstrap interval of the difference, computed by resampling
sections, excludes zero for all four language-model configurations. The class \emph{qhs} to
\emph{q1h} has 200 to 216 pairs per configuration, and the interval excludes zero only for
Llama ZS and Llama 1S.

\begin{table}[t]

\caption{Inter-entry interference over all returned entries (\% of scored pairs). Each row is one
configuration and each column one perturbation class. A cell gives the share of scored pairs in
which at least one other entry of the section changed. For a language model, the upper row compares
the perturbed output with the ten runs of the unperturbed section, and the row ``repeat runs''
compares those ten runs with each other. -- : fewer than 30 scored pairs.}
\label{tab:interference}
\centering
\footnotesize
\setlength{\tabcolsep}{3pt}
\begin{tabular}{lrrrrrrr}
\toprule
  Configuration & Drug-name substitution & Tenfold dose & Dose deletion & Daily to QID & qhs to q1h & Frequency deletion & Route deletion \\
\midrule
cTAKES    & 1.3  & 1.5  & 2.9  & --   & --   & 4.6  & 3.8  \\
Med7      & 0.3  & 0.1  & 0.3  & 1.4  & 1.6  & 1.5  & 1.1  \\
Med7-B    & 7.2  & 0.2  & 0.4  & 0.8  & 2.2  & 1.0  & 0.4  \\
Med7-trf  & 28.6 & 7.5  & 45.0 & 12.3 & 28.7 & 45.6 & 32.5 \\
\midrule
Llama ZS  & 29.0* & 20.8* & 35.9* & 24.5* & 18.4* & 31.1* & 36.0* \\
\quad repeat runs & 15.7 & 15.5 & 15.5 & 14.4 & 11.4 & 14.7 & 15.0 \\
Llama 1S  & 28.6* & 22.1* & 36.3* & 23.9* & 24.0* & 32.5* & 31.1* \\
\quad repeat runs & 17.7 & 18.0 & 18.0 & 17.6 & 17.7 & 17.4 & 16.9 \\
Llama CoT & 24.8* & 18.9* & 30.8* & 21.9* & 17.4  & 27.2* & 26.2* \\
\quad repeat runs & 15.5 & 15.6 & 15.5 & 14.8 & 15.6 & 15.0 & 14.1 \\
Qwen ZS   & 44.9* & 36.2* & 51.8* & 38.5* & 33.1  & 46.9* & 46.1* \\
\quad repeat runs & 30.0 & 30.0 & 30.0 & 29.0 & 29.4 & 29.8 & 29.7 \\
\bottomrule
\end{tabular}

\smallskip
* The 95\% interval of the difference from repeat runs excludes zero.
\end{table}

Table~\ref{tab:intra} gives the intra-entry interference rate by class and configuration. The
rate is the share of scored pairs in which a non-edited field of the edited entry changed. Pairs
whose entry is missing from the perturbed output are not in the table (at most 241 per cell).
For cTAKES, Med7, Med7-B, and Med7-trf the rate is at most 1.1\% for a tenfold dose and for
\emph{qhs} to \emph{q1h}. For a deleted dose it is 0.8\% to 1.6\%, and 3.7\% for Med7-trf. It is
higher in four classes: drug-name substitution (4.5\% to 6.0\%), daily to QID (4.1\% to 4.4\% for Med7,
Med7-B, and Med7-trf), frequency deletion (7.4\% to 9.3\%), and route deletion (7.0\% and 7.8\%
for Med7 and Med7-B, 2.3\% for Med7-trf).

For the language models, the perturbed output is compared with each of the ten runs of the
unperturbed section, and the rate is the average over these comparisons. A comparison in which
the entry is missing from either output is left out (at most 1.5\% of the comparisons of a
cell). The row ``repeat runs'' gives the same rate between the ten unperturbed runs, where no
edit was made. When the route is deleted the rate is 22.7\% to 62.5\%, against 0.7\% to 2.2\%
between repeat runs. For the drug-name substitution, the deleted dose, daily to QID, and the deleted
frequency the rate is above repeat runs in every configuration. For the tenfold dose the rate is
2.4\% to 4.7\%, and repeat runs alone give 1.7\% to 2.7\%. The class \emph{qhs} to \emph{q1h}
has 200 to 216 pairs per configuration and is not interpreted. The changes concentrate in the
distorted pairs. Measured against the reference run, and among the pairs whose entry is returned,
36\% to 50\% of the distorted pairs with a deleted dose (195 to 274 pairs per configuration) also changed another field, against 7\% to 15\% of the directly propagated pairs. For the drug-name substitution the shares are 40\% to 60\% and 8\% to 14\%, on only 20 to 30 distorted pairs per configuration.

\begin{table*}[!t]
\caption{Intra-entry interference (\% of scored pairs). Each row is one configuration and each
column one perturbation class. A cell gives the share of scored pairs in which a field of the
edited entry other than the edited one changed; pairs whose edited entry is missing from the
perturbed output are left out. For a language model, the upper row compares the perturbed output
with the ten runs of the unperturbed section, and the row ``repeat runs'' compares those ten runs
with each other. -- : fewer than 30 scored pairs.}
\label{tab:intra}
\centering
\footnotesize
\begin{tabular}{lrrrrrrr}
\toprule
Configuration & Drug-name substitution & Tenfold dose & Dose deletion & Daily to QID & qhs to q1h & Frequency deletion & Route deletion \\
\midrule
cTAKES    & 5.8 & 0.0 & 0.8 & --  & --  & 7.4  & 0.3 \\
Med7      & 4.8 & 0.2 & 1.6 & 4.1 & 0.0 & 9.3  & 7.0 \\
Med7-B    & 4.5 & 0.2 & 1.5 & 4.4 & 1.1 & 9.0  & 7.8 \\
Med7-trf  & 6.0 & 0.6 & 3.7 & 4.2 & 0.0 & 8.4 & 2.3 \\
\midrule
Llama ZS  & 9.5*  & 4.2* & 15.6* & 8.0* & 3.7* & 9.0*  & 62.5* \\
\quad repeat runs & 2.9 & 2.3 & 2.2 & 1.4 & 0.3 & 1.6 & 1.5 \\
Llama 1S  & 8.7*  & 2.8* & 12.6* & 7.9* & 6.2* & 7.8*  & 42.4* \\
\quad repeat runs & 2.0 & 1.8 & 1.7 & 1.0 & 1.2 & 1.5 & 0.8 \\
Llama CoT & 8.0*  & 2.4* & 8.2*  & 4.5* & 1.3  & 7.6*  & 22.7* \\
\quad repeat runs & 1.9 & 1.7 & 1.7 & 1.2 & 0.7 & 1.6 & 0.7 \\
Qwen ZS   & 13.6* & 4.7* & 13.1* & 6.3* & 2.8* & 17.3* & 35.5* \\
\quad repeat runs & 3.6 & 2.7 & 2.7 & 2.7 & 1.2 & 3.0 & 2.2 \\
\bottomrule
\end{tabular}

\smallskip
* The 95\% interval of the difference from repeat runs excludes zero.
\end{table*}

\subsubsection{Total error}

\begin{figure*}[!t]
\centering
\includegraphics[width=\textwidth]{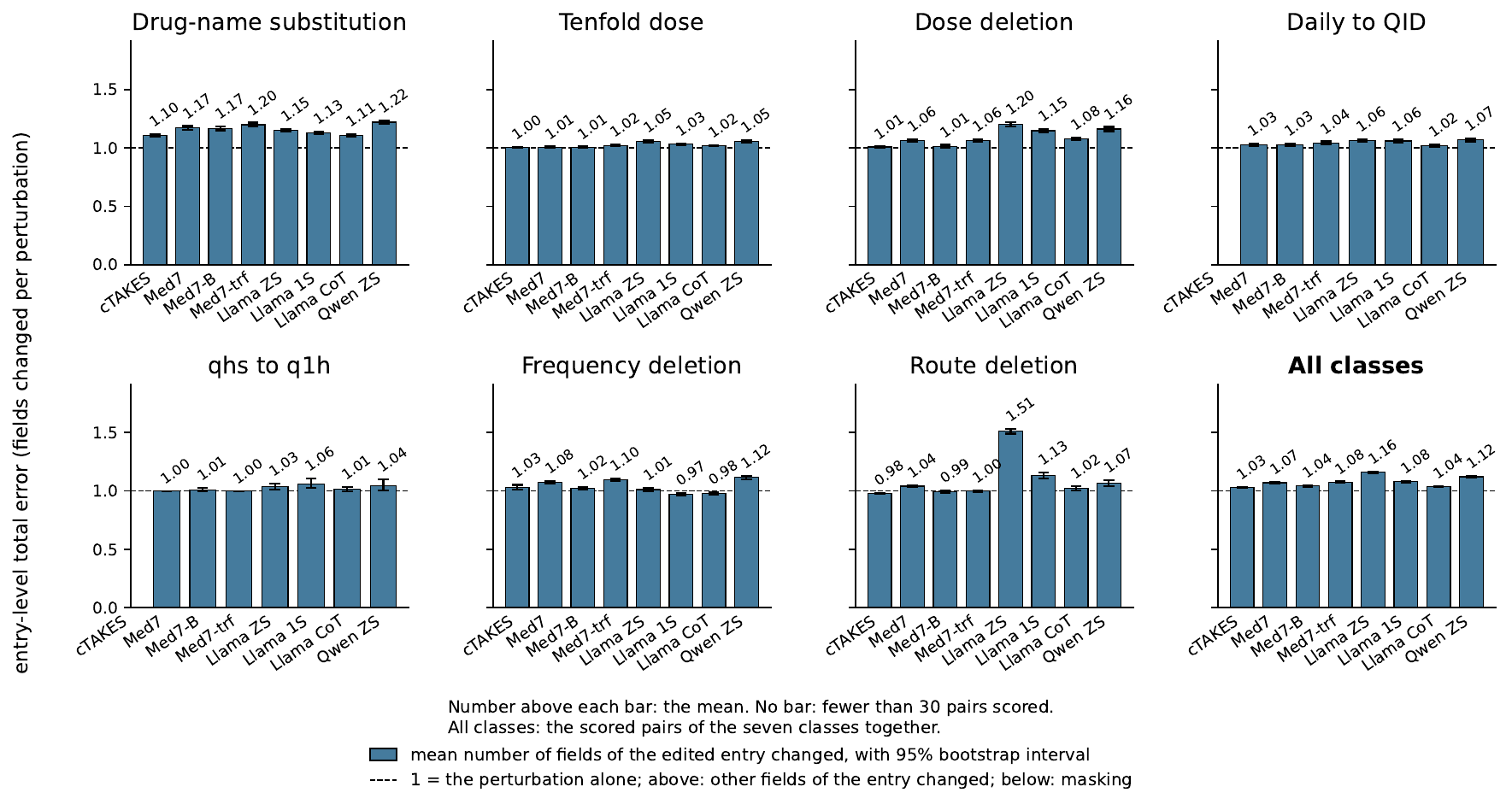}
\Description{Eight panels, one per perturbation class and a last panel for all classes. Each
panel shows, for every configuration, the mean number of fields of the edited entry that changed
per perturbation, with a 95 percent interval. A dashed horizontal line at 1 marks the perturbation
alone.}
\caption{Entry-level total error by class and configuration: mean number of fields of the edited entry changed per perturbation, with 95\% bootstrap intervals. The dashed line at 1 is the perturbation alone. The last panel takes the scored pairs of the seven classes together.}
\label{fig:total}
\end{figure*}

Fig.~\ref{fig:total} gives the entry-level total error, $a + b$. It is the number of fields of
the edited entry that differ between the perturbed and the reference output, averaged over
scored pairs. A value of 1 means the perturbation reached the output and nothing else in the
entry changed. A value below 1 means some pairs were masked, since a masked pair counts 0. A
pair whose edited entry is missing from the perturbed output has a total error of 4; these are
the pairs left out of Table~\ref{tab:intra}. Every value lies between 0.97 and 1.22, except
one. For a deleted route, Llama ZS reaches 1.51, and 60.8\% of its pairs have exactly two
changed fields. In these pairs the frequency moves into the route field and the frequency field
empties. Llama 1S (1.13), Llama CoT (1.02), and Qwen ZS (1.07) do this less often. Across
all classes the share of pairs with three or more changed fields is at most 7.6\%, and above
5\% only for a drug-name substitution and a deleted dose. For the language models, repeat runs alone change
0.01 to 0.07 fields of the edited entry. When the perturbed output is compared with all ten
unperturbed runs instead of the reference run, every value changes by at most 0.02. The last
panel of Fig.~\ref{fig:total} takes the scored pairs of the seven classes together. There the
entry-level total error is 1.03 to 1.16: lowest for cTAKES (1.03) and highest for Llama ZS
(1.16). As in the last panel of Fig.~\ref{fig:fates-b}, a class counts in proportion to its
scored pairs, and the interval understates the uncertainty.

Table~\ref{tab:totalsection} gives the third part of the total error, $c$: the number of fields
changed in other entries per perturbation, over all returned entries. cTAKES and Med7 change
at most 0.07 fields. Med7-B changes at most 0.09, except for the drug-name substitution (0.28). Med7-trf
changes 0.28 to 2.30 fields. For a deleted dose it changes 1.06 fields in the edited entry and
2.24 in other entries, a section-level total error of 3.3. The language models have two rows,
as in Table~\ref{tab:interference}. Repeat runs alone change 0.44 to 0.92 fields in other
entries for the Llama configurations and 2.11 to 2.39 for Qwen ZS. The edit adds to these. For the
drug-name substitution and the three deletion classes it adds 0.45 to 1.84 fields for the Llama
configurations and 0.76 to 1.29 for Qwen ZS, and the 95\% bootstrap interval of the difference
excludes zero in every case. The largest addition is for Llama ZS with a deleted route:
1.51 fields change in the edited entry and 1.84 more in other entries than between repeat runs.
For the tenfold dose and daily to QID, Llama ZS and Llama 1S add 0.24 to 0.53 fields and
Qwen ZS 0.25 and 0.42. Llama CoT adds 0.24 fields for daily to QID, and for the tenfold dose
its interval includes zero. The class \emph{qhs} to
\emph{q1h} has 200 to 216 pairs per configuration, and its differences are not interpreted.

\begin{table}[t]
\caption{Fields changed in other entries per perturbation, over all returned entries. Each row
is one configuration and each column one perturbation class. A cell gives the mean number of
fields that changed in the other entries of the section per scored pair, which is part $c$ of the
total error. For a language model, the upper row compares the perturbed output with the ten runs
of the unperturbed section, and the row ``repeat runs'' compares those ten runs with each other.
-- : fewer than 30 scored pairs.}
\label{tab:totalsection}
\centering
\footnotesize
\setlength{\tabcolsep}{2.5pt}

\begin{tabular}{lrrrrrrr}
\toprule
Configuration & Drug-name substitution & Tenfold dose & Dose deletion & Daily to QID & qhs to q1h & Frequency deletion & Route deletion \\
\midrule
cTAKES    & 0.02 & 0.02 & 0.07 & --   & --   & 0.05 & 0.06 \\
Med7      & 0.01 & 0.00 & 0.01 & 0.02 & 0.07 & 0.03 & 0.01 \\
Med7-B    & 0.28 & 0.01 & 0.01 & 0.02 & 0.09 & 0.03 & 0.01 \\
Med7-trf  & 1.29 & 0.28 & 2.24 & 0.44 & 1.01 & 2.30 & 1.35 \\
\midrule
Llama ZS  & 1.73* & 1.08* & 1.96* & 1.16* & 0.71* & 1.50* & 2.55* \\
\quad repeat runs & 0.78 & 0.78 & 0.78 & 0.63 & 0.44 & 0.68 & 0.71 \\
Llama 1S  & 1.45* & 1.02* & 1.69* & 1.07* & 0.95* & 1.47* & 1.60* \\
\quad repeat runs & 0.77 & 0.78 & 0.78 & 0.74 & 0.64 & 0.74 & 0.69 \\
Llama CoT & 1.34* & 1.01  & 1.48* & 1.07* & 0.75  & 1.27* & 1.37* \\
\quad repeat runs & 0.83 & 0.92 & 0.86 & 0.83 & 0.84 & 0.83 & 0.80 \\
Qwen ZS   & 3.38* & 2.64* & 3.68* & 2.77* & 2.31  & 3.05* & 2.95* \\
\quad repeat runs & 2.39 & 2.38 & 2.38 & 2.34 & 2.14 & 2.30 & 2.11 \\
\bottomrule
\end{tabular}

\smallskip
* The 95\% interval of the difference from repeat runs excludes zero.
\end{table}

\subsection{\mshi{Comparison of drug-name similarity measures}}
\label{sec:similarity}
\mshi{The composite score that selects the substitute names (Section~\ref{sec:classes}) was chosen after a comparison with six other measures.} Seven measures were compared with the evaluation of Kondrak and Dorr~\cite{kondrak2006confusable},
with the ISMP list as the reference. They are the composite, its three components (BI-SIM,
normalized edit similarity, and ALINE), Editex~\cite{zobel1996phonetic}, the Dice coefficient on
character bigrams~\cite{dice1945measures}, and Jaro--Winkler~\cite{winkler1990string}. Each ISMP
name pair whose two names resolve in RxNorm gives one query. The first name is compared with all
19{,}807 ingredient and brand names, and the rank of its documented partner is recorded. This gives
599 queries from 424 names. ALINE is applied to the names as spelled. Several names often have the
same score as the partner: in 88 to 525 of the 599 queries, depending on the measure. The rank of
the partner is then not unique. Every reported value is the expectation over all orders of the
tied names.

\begin{table}[!b]
\caption{Recovery of the ISMP confusion partner by similarity measure. Each row is one measure.
The first four value columns are computed over the 599 queries: recall at ranks 1, 5, and 10 (R@1,
R@5, R@10) and the mean reciprocal rank (MRR). The last column is recall at rank 10 averaged per
name, over the 424 names. Rows are ordered by MRR, from highest to lowest.}
\label{tab:lasa}
\centering
\footnotesize
\setlength{\tabcolsep}{3.5pt}
\begin{tabular}{@{}lrrrrr@{}}
\toprule
 & \multicolumn{4}{c}{Per query} & Per name \\
Measure & R@1 & R@5 & R@10 & MRR & R@10 \\
\midrule
Composite             & 0.129 & 0.316 & 0.414 & 0.225 & 0.442 \\
Editex                & 0.129 & 0.315 & 0.416 & 0.224 & 0.436 \\
BI-SIM                & 0.128 & 0.312 & 0.410 & 0.219 & 0.430 \\
Dice (bigrams)        & 0.123 & 0.306 & 0.380 & 0.212 & 0.393 \\
Normalized edit sim.  & 0.117 & 0.296 & 0.397 & 0.206 & 0.416 \\
Jaro--Winkler         & 0.108 & 0.308 & 0.404 & 0.205 & 0.431 \\
ALINE (on spelling)   & 0.088 & 0.229 & 0.283 & 0.161 & 0.291 \\
\bottomrule
\end{tabular}
\end{table}

Table~\ref{tab:lasa} reports recall at ranks 1, 5, and 10 and the mean
reciprocal rank (MRR) over the queries. It also reports recall at rank 10
averaged per name, as Kondrak and Dorr do. The composite and Editex \mshi{have nearly the same MRR}.
The MRR of Editex is lower by 0.001, with a 95\% interval of the difference of
$[-0.016, 0.014]$ from a paired bootstrap of 2{,}000 replicates that
resample queries~\cite{efron1993bootstrap}. BI-SIM, Dice, and
Jaro--Winkler are not distinguishable from the composite in MRR either.
Normalized edit similarity and ALINE are lower, and their intervals exclude zero ($[-0.034, -0.006]$
and $[-0.081, -0.048]$). The MRR of Jaro--Winkler is close to that of normalized edit similarity, but
its interval is wider and includes zero ($[-0.045, 0.005]$).
At rank 20 of the per-name evaluation, Editex (0.544) and
BI-SIM (0.535) are above the composite (0.507). The composite is used
because no measure exceeds it in MRR and it has the published form of the
FDA's program. The benchmark ranks over all 19{,}807 names to test the
measures. The substitution itself draws only from names in clinical use,
so that every substitute is a drug a clinician could prescribe.

The composite was then compared with the FDA's POCA tool
(\url{https://poca-public.fda.gov}, Public Version 2.22.0, accessed
November 15, 2026). All 343 distinct name pairs, an original name and its
substitute from tiers~2 and~3, were scored. The original name was entered
as the proposed name, because the tool's score depends on the order of
the two names. The tool scores 330 of the 343 name pairs (96.2\%) at 55 or
above, its threshold for moderate similarity; the 95\% Wilson interval is
93.6\% to 97.8\%~\cite{wilson1927probable}. All 88 name pairs of tier~2 reach
55, and 83 of them reach 70. The 13 name pairs below 55 cover 270 of the
3{,}837 entries of these tiers; 186 of them are one name pair, docusate and
docetaxel, which the tool scores at 52. The two scores differ by 3.9
points on average, and the tool's score is the higher one in 80.5\% of the name pairs.

\section{Discussion and Limitations}
\label{sec:discussion}

\subsection{Extraction is not a safety net}

The three fates of this study, masked, propagated directly, and propagated with distortion,
describe what a perturbation does to the structured medication fields. They do not describe
whether an extraction system recognizes an error in the note. A masked pair may look like a
correction, but it means only that the perturbed output equals the reference output in the
edited field. The configurations extract the fields in different ways. cTAKES uses a dictionary
and fixed patterns, Med7 is a tagger trained on annotated text, and the language models generate
the field values. \mshi{In all eight configurations, 98.6\% to 100\% of drug-name substitutions and tenfold doses were nevertheless propagated, directly or with distortion. When cTAKES recognized the substituted name, it assigned the name its own RxNorm code,} so linking a name to a terminology code did not keep the substituted name out of the
structured output\mshi{; in 64 of the 79 pairs in which the entry was missing from the cTAKES output, the substitute name was missing from its dictionary (Section~\ref{sec:limitations})}. Masking was rare except for deleted values. There it occurred when the entry
repeated the value or the configuration supplied a default. Masking therefore reflects
information available elsewhere in the entry or a default of the configuration, not recognition
of an error. No configuration was designed or prompted to check the note, so the study does not
show whether any of them could detect such errors if asked. The results show that extraction
alone does not remove these errors. A check has to be added, on the note before extraction or on
the structured output against an independent source.

\subsection{Masking by extraction method}

The masking of a deleted value depends on how much of the entry a configuration reads. An entry
can repeat a value on its pharmacy \emph{RX} line or on the numbered line itself. cTAKES
searches the sentence that contains the medication, and that sentence ends at the line break, so
cTAKES does not read the \emph{RX} line. Med7 grouped by line does not read it either. Both
seldom return a deleted value that the \emph{RX} line repeats, \mshi{but} they return one that the
numbered line states twice. Med7 grouped by entry block reads the \emph{RX} line and often
returns the value. The tagger is the same in both groupings, so the difference comes from the
grouping rule alone. The language models receive the whole section, and they also return the
repeated value often. When the entry holds no copy, a deleted dose is seldom recovered by any
configuration. The language models have a second source of masking: they fill a missing route
with a default such as \emph{PO}. The other configurations seldom do so. The default is not read
from the note.

\subsection{Distortion and interference by extraction method}
The categories also differ in what they return when a value is
missing from the note. cTAKES and the Med7 configurations recognize
each field by its own pattern or label, and when the route is
deleted they leave the field empty. The language models generate
the field values. They often place the frequency of the entry in
the empty route field, so one deleted field gives two changed
fields. The one-shot and chain-of-thought prompts reduced this
behavior \mshi{but} did not remove it.

The language models also change fields that the edit did not touch. Part of this is run-to-run
variation: their output for the same unperturbed section differs between runs with identical
settings. The edit adds to it, in the other fields of the edited entry and in the other entries
of the section. Such changes are not limited to the language models. Med7-trf is deterministic,
and one edit still often changes other entries, while Med7 with the same grouping rule seldom
does. The two configurations differ only in the model. The transformer encoder of Med7-trf
represents each word using the other words in a window of 256 tokens, the setting in the
configuration distributed with the model, so a change to one entry can change the representation of words in other entries. The tagger of Med7 is a convolutional network~\cite{kormilitzin2021med7}, which represents a word using only a limited number of neighboring words. Deterministic output is therefore not sufficient to keep an error within the entry that contains it.

Table~\ref{tab:strengths} summarizes the strengths and weaknesses that each category of
extraction system showed in this study. They hold for these configurations and this note format,
and are not a general ranking of the categories.

\begin{table*}[!b]
\caption{Strengths and weaknesses by category of extraction system, as observed in this study.
Each row is one category. From left to right: the category, its configurations, the strengths
observed, and the weaknesses observed. They hold for these configurations and this note
format.}
\label{tab:strengths}
\centering
\footnotesize
\begin{tabular}{p{0.13\textwidth}p{0.16\textwidth}p{0.30\textwidth}p{0.31\textwidth}}
\toprule
Category & Configurations & Strengths & Weaknesses \\
\midrule
Dictionary-based & cTAKES & Finds almost every drug name; rarely changes other entries & Rarely attaches a frequency; gives a substituted drug name a valid code \\
\addlinespace
Supervised tagger & Med7, Med7-B, Med7-trf & Stable output; block grouping recovers a \mbox{repeated} dose & Depends on the grouping rule; transformer \mbox{encoder} changes other entries \\
\addlinespace
Language model & Llama ZS, 1S, and CoT; Qwen ZS & Extracts the most entries; recovers values \mbox{repeated} on other lines & Moves values between fields; output varies from run to run \\
\bottomrule
\end{tabular}
\end{table*}

\subsection{Limitations}
\label{sec:limitations}

Only entries that a configuration extracts correctly from the unperturbed section are scored.
Correct extraction and masking are decided after normalization: differences that normalization
removes are not counted as changes, and wording variants that it does not reduce are. cTAKES
attaches a frequency to few entries in this note format, so its frequency classes are mostly
unscored. Its dictionary lacks some substitute names: 64 of the 79 drug-name substitution pairs in
which the entry is missing from the cTAKES output involve five such names. Med7 and the encoder of
Med7-trf were trained on MIMIC-III text (Section~\ref{sec:pipelines}), which comes from the same
hospital as the notes of this study, so their training data may include notes of the same patients.
The training corpus of Med7-trf itself is not described.

Run-to-run variation was measured on ten runs of each unperturbed section, \mshi{all under concurrent inference}, and each perturbed
section was extracted once. For the language models, the fate was scored against each of the ten
runs, and the interference rates and the total error were compared with repeat runs. A pair of a
language model is scored when the first of the ten runs extracted the entry correctly. The other
nine runs carry no such condition, so a comparison with them can include a run that did not
extract the entry correctly.

The medication labels were generated by the dataset authors using a language-model pipeline
rather than human annotation~\cite{goel2023medlabels}, and they were audited by hand only on a
sample (Section~\ref{sec:preparation}). Five inclusion criteria selected the
entries of this study, so entries excluded by those criteria are not represented. A pair is
scored only when the reference output agrees with the label. An incorrect label generally causes
that pair to be excluded rather than assigned an incorrect fate. The fate assignment itself was
validated by a single annotator, an author of this study, and the token-level criterion was
defined after that validation, from its disagreements.

The perturbations are limited to a single edit in one medication field, whereas errors in real
clinical documentation may occur together. Every perturbation counts the same. The study
measures propagation, not clinical impact, and a severity weighting would need clinical input.
The notes are discharge summaries from one hospital, in English. They were written by clinicians,
not generated by an AI scribe. The perturbations are of the kinds documented in AI-scribe and
dictation output, but the layout of scribe-generated notes may differ from these summaries. The
two note templates are those of MIMIC-IV discharge summaries, and masking depends on where a note
repeats a value, so masking rates would differ for other note formats. The daily-to-QID and
tenfold-dose classes reproduce the outcome of documented error mechanisms rather than their exact
triggering conditions. The study examines only one stage of the documentation process, from the
medication list of a clinical note to the structured medication output. Error propagation from
the spoken encounter to the clinical note is outside its scope. All models ran on one local
machine, so that no note text left it. This limited the evaluation to open-weight language models
that fit on one GPU: two models of similar size, both with 4-bit quantization. The prompts were
not tuned on these notes. Under the chain-of-thought prompt, the result for 35.2\% of the inputs comes from the second request (Section~\ref{sec:configurations}), so that configuration combines replies with and
without a completed justification.

\section{Conclusion}
\label{sec:conclusion}

This study asked whether medication extraction detects, corrects, or passes on errors already
present in a clinical note. On seven classes of documented medication errors, introduced one at a
time into 7{,}256 medication entries and processed by eight extraction configurations, the answer is that extraction passes them on. No configuration was asked to check the note, and
extraction alone did not remove the errors. Drug-name substitutions and tenfold dose errors \mshi{were propagated, directly or with distortion,} in 98.6\% to 100\% of cases. Deleted fields were masked in
0.3\% to 31\% of cases. For these, masking above 3\% occurred only where the entry repeated the missing
value or the configuration supplied a default, so it reflects redundancy in the note, not
recognition of an error. \mshi{The language models and Med7 grouped by entry block masked more deletions because they read the whole entry, but the language models} also shifted values between fields of the same entry. \mshi{Under concurrent inference, their} outputs varied from run to run:
the output was identical in all ten runs for 53.9\% to 70.1\% of sections. In the language
models, a drug-name substitution or a field deletion changed other entries more often than repeat
runs did, by 9.4 to 21.8 percentage points. A deterministic tagger with a transformer encoder,
Med7-trf, changed another entry in 7.5\% to 45.6\% of pairs, so interference with other entries
does not require run-to-run variation. Preventing these errors from reaching the structured
output therefore requires detecting them in the note before extraction or validating the
extracted output against an independent source.

\section*{\mshi{Data Availability}}
\mshi{The discharge summaries, the structured record, and the medication labels are distributed by PhysioNet to credentialed users under a data use agreement~\cite{johnson2023mimicnote, bennett2024mimicfhir, johnson2024mimiciv31, goel2023medlabels}. The agreement does not allow the note text to be reproduced, so the examples shown in the figures are constructed.}

\bibliographystyle{ACM-Reference-Format}
\bibliography{reference}

\end{document}